\documentclass[11pt]{article}

\usepackage[final]{acl}
\usepackage{times}
\usepackage{latexsym}

\usepackage[T1]{fontenc}

\usepackage[utf8]{inputenc}

\usepackage{microtype}

\usepackage{inconsolata}

\usepackage{graphicx}

\usepackage{amsmath}
\usepackage{amssymb}
\usepackage{mathtools}
\usepackage{amsthm}

\usepackage[capitalize,noabbrev]{cleveref}

\theoremstyle{plain}

\theoremstyle{definition}

\theoremstyle{remark}

\usepackage[textsize=tiny]{todonotes}

\usepackage{booktabs}
\usepackage{url}
\usepackage{graphicx}
\usepackage{threeparttable}
\usepackage{multirow}
\usepackage{amsmath}
\usepackage{adjustbox}
\usepackage{float}
\usepackage{array}    
\usepackage{makecell} 
\usepackage{siunitx}  
\usepackage{graphicx} 
\usepackage{caption}  
\usepackage{arydshln}
\definecolor{mybrown}{RGB}{128,64,0}
\usepackage{isabelle,isabellesym}

\usepackage{enumitem}
\setlist[enumerate]{leftmargin=13pt, labelsep=4pt}
\usepackage[utf8]{inputenc}
\usepackage[T1]{fontenc}
\definecolor{lightred}{RGB}{200, 100, 50}
\usepackage{tcolorbox}
\usepackage{listings}
\usepackage{color}
\usepackage{setspace}

\definecolor{isarblue}{HTML}{006699}
\definecolor{isarfaintblue}{rgb}{0.0, 0.75, 1.0}
\definecolor{isargreen}{HTML}{009966}
\definecolor{red}{HTML}{990000}
\definecolor{patriarch}{rgb}{0.5, 0.0, 0.5}

\lstdefinelanguage{isabelle}{
	keywords=[1]{type_synonym,datatype,fun,abbreviation,definition,proof,lemma,theorem,qed,corollary,have,hence,also,finally,ultimately,moreover,using,\{},
	keywordstyle=[1]\bfseries\color{isarblue},
	keywords=[2]{where,assumes,shows,fixes,and,by,sorry},
	keywordstyle=[2]\bfseries\color{isargreen},
	keywords=[3]{if,then,else,case,SOME,let,in,O},
	keywordstyle=[3]\color{isarblue},
	keywords=[4]{ATP},
	keywordstyle=[4]\it\color{patriarch},
	keywords=[5]{show,assume,obtain},
	keywordstyle=[5]\bfseries\color{isarfaintblue},
}

\lstdefinestyle{isabelle}{
	language=isabelle,
	escapeinside={\&}{&},
	columns=fixed,
	extendedchars,
	basewidth={0.5em,0.45em},
	basicstyle=\singlespacing\ttfamily\small,
	mathescape,
	morecomment=[s][\bfseries\color{red}]{(*}{*)},
	morecomment=[l][\bfseries]{####},
}

\definecolor{lightgreen}{RGB}{40, 160, 40}
\definecolor{lightred}{RGB}{200, 60, 60}

\title{Improving Autoformalization Using Direct Dependency Retrieval}

\author{
	Shaoqi Wang$^\dagger$ \quad Lu Yu$^\ddagger$ \quad Siwei Lou$^\dagger$ \quad Feng Yan$^*$ \\\quad \textbf{Chunjie Yang}$^\dagger$ \quad \textbf{Qing Cui}$^\ddagger$ \quad \textbf{Jun Zhou}$^\ddagger$    \vspace{5pt}\\
	$^\dagger$ Zhejiang University, $^\ddagger$ Ant Group, $^*$ Hong Kong Polytechnic University
}

\begin{document}
\maketitle
\begin{abstract}
Statement autoformalization, a crucial first step in formal verification, aims to transform informal descriptions of math problems into machine-verifiable formal representations but remains a significant challenge. 
The core difficulty lies in the fact that existing language models often suffer from a lack of contextual awareness, leading to hallucination of formal dependencies such as definitions and theorems. 
Current dependency retrieval approaches exhibit poor precision and recall, and lack the scalability to leverage ever-growing public datasets. To bridge this gap, we propose a novel retrieval-augmented framework based on Direct Dependency Retrieval (DDR). 
DDR directly generates candidate formal dependencies from natural-language mathematical descriptions and verifies their existence in the formal library via an efficient Suffix Array Check (SAC). 
Built on a SAC-constructed dependency retrieval dataset of over 500,000 samples, a high-precision DDR model is fine-tuned and shown to significantly outperform state-of-the-art methods in both retrieval precision and recall, leading to superior advantage in the autoformalization tasks.
SAC also contributes in assessing formalization difficulty and enabling explicit quantification of the hallucination in In-Context Learning (ICL).
\end{abstract}

\section{Introduction}

Theorem provers, such as Lean \citep{moura2021lean}, Coq \citep{bertot2013interactive}, and Isabelle \citep{nipkow2002isabelle}, are symbolic systems designed for formal verification whose soundness and completeness are proven. Grounded in specific foundational logics, these systems employ proprietary formal languages to encode mathematical statements, definitions, and proofs. The logical kernel within each system can then verify the validity of these formalized objects with complete rigor \citep{nawaz2019survey}. However, this process eschews the flexibility of natural language, demanding that users not only master the strict formal syntax but also possess a profound understanding of the semantic mapping between mathematical ideas and their formal representations. Consequently, the effective utilization of these theorem provers presents a formidable challenge to researchers \citep{li2024survey}.

Statement autoformalization refers to the process of automatically translating mathematical statements from natural language into a formal language comprehensible to theorem provers \citep{liu2025rethinking}. As a foundational step in the formal verification, the correctness of its output is paramount. It determines whether the proof target is precisely aligned with the intent of the original statement. 
Beyond this core function, statement autoformalization exhibits significant potential for broader applications and innovation. These applications include synthesizing training data for neural theorem provers \citep{lin2025goedel}, automated program verification \citep{lin2024fvel}, and curating and filtering informal reasoning datasets \citep{zhang2025deeptheorem}.
, and evaluating and enhancing the informal reasoning capabilities of LLMs (e.g., through techniques like rejection sampling \citep{zhou2024don}).

Current mainstream research in autoformalization primarily encompasses three direction. First, constructing larger and more challenging datasets for Supervised Fine-Tuning (SFT) \citep{cao2025informal,peng2025criticlean,yu2025formalmath,miranda2025veribench,kimina_prover_2025,ying2024lean,zheng2021minif2f,azerbayev2023proofnet}; second, encoding and indexing the libraries of theorem provers to support retrieval \citep{liu2025rethinking,yang2023leandojo,wang2025aria}; and third, employing methods like ICL to enable Large Language Models (LLMs) to directly translate informal statements into formal code \citep{wu2022autoformalization,zhang2024consistent}.

\begin{figure*}
	\centering
	\includegraphics[width=.9\textwidth]{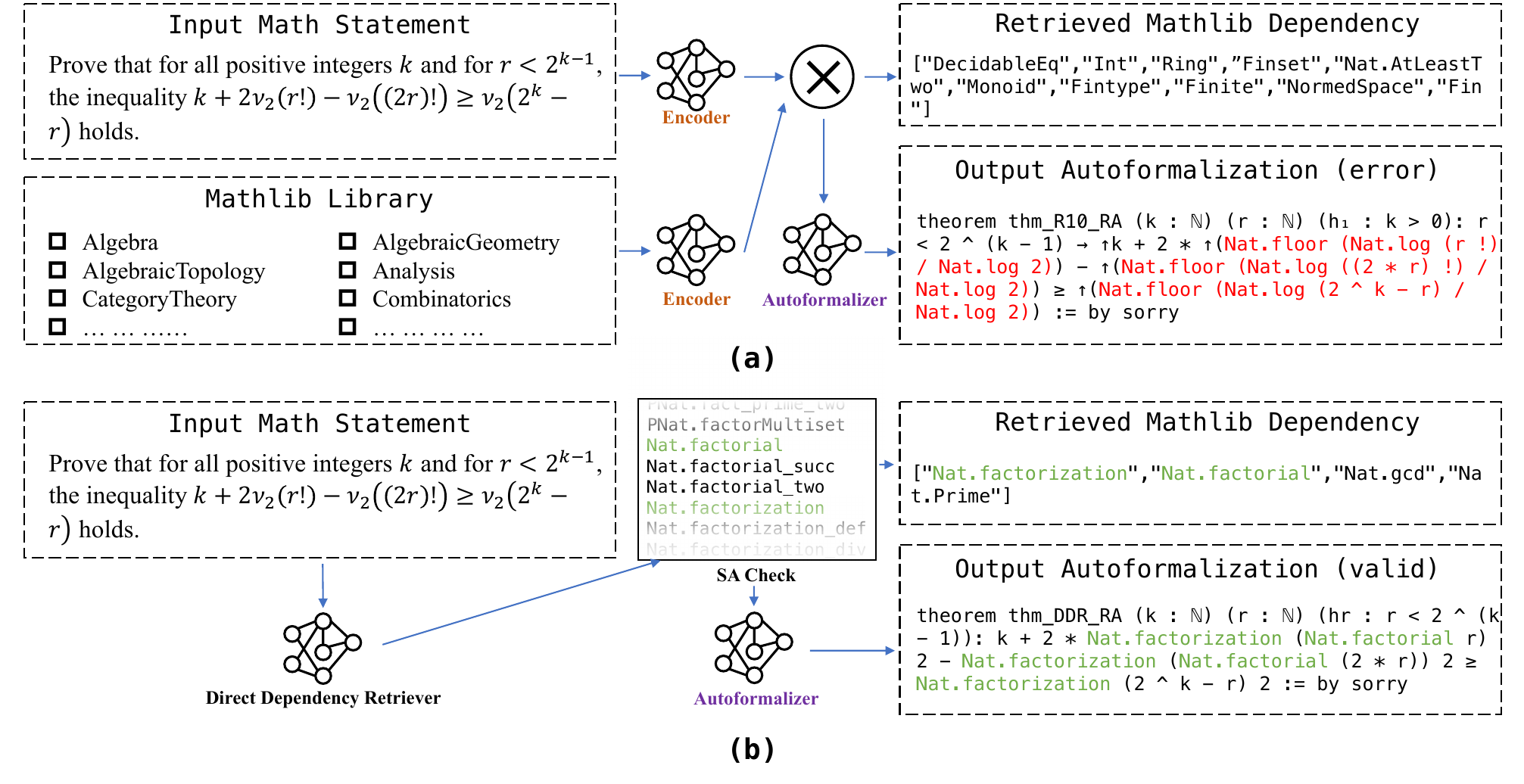}
	\caption{{Comparison of two paradigms for formal dependency retrieval. The \textit{v}$_2(\cdot)$ in input indicates 2-adic valuation. (a) Select-based retrieval using embedding cosine similarity: both the input statement and library items are encoded into embeddings, and the top-k items with highest cosine similarity are retrieved as dependencies. (b) DDR generates potential dependencies based on context, followed by SAC verification step.}}
	\label{fig:example-0}
\end{figure*}

However, these approaches face two critical limitations. First, relying solely on generative models to directly synthesize formal code often overlooks the existing definitions and lemmas within the formal library \citep{yang2023leandojo,zhang2024consistent}. This ignorance is a primary reason for the model to hallucinate non-existent formal objects or generate syntactically incorrect code \citep{wu2022autoformalization,liu2025rethinking}. 
Results presented in Table \ref{tab:dependency-retrieval-hall} demonstrate that even with a two-stage ICL approach, where dependencies are retrieved prior to the main task, the problem of hallucination in model outputs remains significant.
Second, the retrieval strategies for library dependencies in existing methods are rudimentary, yielding results with weak semantic relevance to the informal statement \citep{azerbayev2023proofnet,yang2023leandojo}. 
The libraries of mainstream theorem provers are vast and continually growing; for instance, Lean 4's mathlib contains over 240,000 entries, including many with similar descriptions and usage. 
Simple retrieval methods struggle to achieve both high efficiency and precision. This challenge is confirmed by the experiments in \citet{liu2025rethinking} as well as our own results presented in Table \ref{tab:dependency-retrieval-icl}.

To address the aforementioned challenges, we propose a novel dependency retrieval method DDR, and the corresponding framework for statement autoformalization. 
In this context, dependency retrieval aims to identify relevant formal objects, such as definitions and theorems, from a theorem prover's library based on an informal mathematical statement \citep{liu2025rethinking}. 
DDR leverages the contextual understanding of LLMs to directly generate potential formal dependencies. 
To empower the LLM with this capability, we have designed SAC for both data generation and dependency verification. 
Specifically, this process begins by constructing an efficient database from the theorem prover's library using a suffix array. 
Subsequently, leveraging existing informal-formal statement pairs from public datasets, we extract candidate dependencies from the formal code and verify them against the database via the efficient binary search. 
This produces high-quality labels for fine-tuning the DDR model. 
The same verification mechanism is reused at inference time to validate the dependencies generated by the DDR model. 
Experimental results demonstrate that DDR significantly improves both recall and precision in dependency retrieval, leading to superior advantage in the autoformalization tasks.

Our main contributions are summarized as follows:
\begin{enumerate}
	\item We propose DDR as a novel generation-plus-verification dependency retrieval paradigm in autoformalization tasks, and SAC as a flexible dependency verification method.
	\item We provide a quantitative analysis of ICL for dependency retrieval, revealing its inherent and severe hallucination problems.
	\item We conduct comprehensive experiments on statement autoformalization. The results validate that DDR can serve as a plug-and-play module to enhance the performance of existing autoformalizers and ICL methods.
    \item The proposed SAC provides a potential sound mechanism for assessing formalization difficulty, overcomes the inherent subjectivity and stochasticity of LLM-based graders.
	\item We have constructed and open-sourced a library dependency dataset with over 500,000 samples. The methodology for constructing this dataset is generalizable and can be applied to other formal corpora, fostering more diverse research in statement autoformalization.
\end{enumerate}

\section{Related Work}

Formal mathematics aims to translate mathematical statements and proofs into machine-verifiable formal languages based on rigorous logical calculus. In recent years, with the advancement of deep learning, this field has become an active area of research, primarily encompassing two core branches: autoformalization \citep{murphy2024autoformalizing,liu2025rethinking,poiroux2024improving,azerbayev2023proofnet,wang2020exploration}, and automated theorem proving \citep{lin2025goedel,kimina_prover_2025,chen2025seed,xin2024deepseek,xin2024deepseekv15,ren2025deepseek,liu2025proofaug}.

\begin{figure}
	\centering
	\includegraphics[width=.45\textwidth]{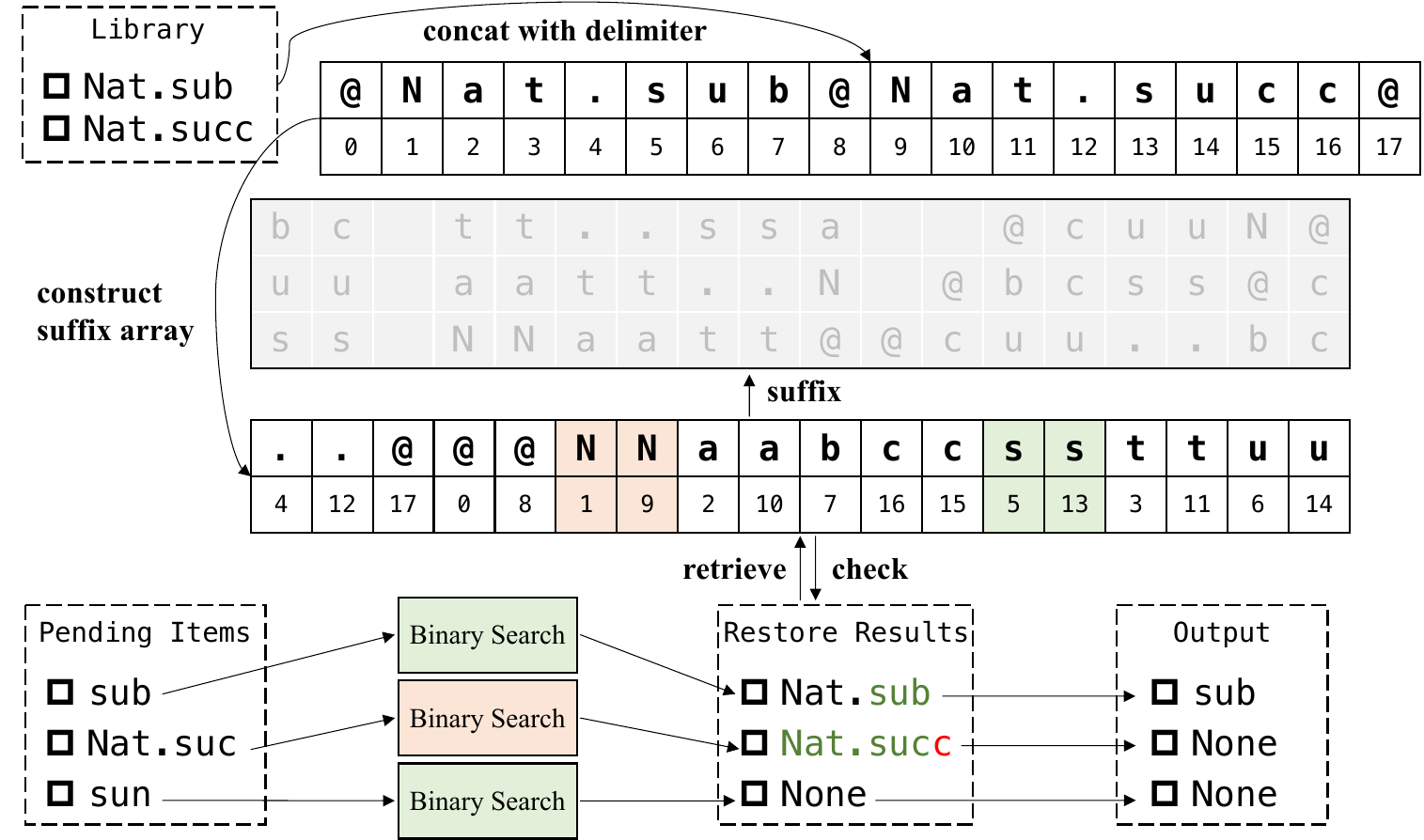}
	\caption{{Schematic diagram of SAC process. All library items are concatenated into a single string with delimiters, followed by suffix array construction. Each pending item is then queried via binary search to determine its match status (exact, partial, or none). This mechanism supports both dataset construction and DDR output verification.}}
	\label{fig:example-1}
\end{figure}

Automated theorem proving is dedicated to automatically generating proof scripts for given formal goals. The central challenge lies in designing efficient proof search strategies \citep{liu2025proofaug}. Current cutting-edge research in this direction focuses on combining LLMs with reinforcement learning to emulate the heuristic reasoning processes of human mathematicians in exploring the proof space \citep{chen2025seed}, with the future prospect of assisting in the discovery of new theorems. Prominent examples of such systems include Seed-Prover \citep{chen2025seed} and DeepSeek-Prover \citep{xin2024deepseek,xin2024deepseekv15,ren2025deepseek}.
In contrast, autoformalization targets the automatic translation of informal mathematical content, typically expressed in natural language, into a formal language. Research in this area concentrates on establishing reliable evaluation methodologies to assess the semantic fidelity of the translation \citep{liu2025rethinking,poiroux2024improving} and on developing higher-performance autoformalization models \citep{liu2025rethinking,poiroux2024improving}.
This paper focuses on the task of statement autoformalization. We propose a method to mitigate model hallucination and enhance model performance by offering better dependency retrieval results in the RAG framework.

Progress in autoformalization research has led to the creation of large-scale informal-to-formal parallel corpora \citep{mahdavi2025leveraging}. These corpora, such as FormalMath \citep{yu2025formalmath}, CriticBench \citep{peng2025criticlean}, and FineLeanCorpus \citep{peng2025criticlean}, have expanded in scale from a few hundred samples to tens of thousands and are primarily generated through rule-based methods and by leveraging the ICL capabilities of LLMs \citep{azerbayev2023proofnet,gao2024omni}. 
However, current autoformalization research has not effectively utilized this data to discover the intricate connection between informal mathematical statements and their required library dependencies. 
To address this gap, we propose DDR, a method designed to enhance a model's ability to infer and utilize library dependencies during statement formalization.

Research on evaluating the fidelity of autoformalization has explored various methods, such as manual expert review \citep{wang2018first,peng2025criticlean},  LLM review \citep{yang2023leandojo,sidhu2024evaluation}, auxiliary metrics like Perplexity \citep{wang2018first}, BLEU \citep{papineni2002bleu,wang2018first}, and type checking \citep{poiroux2024improving} within a theorem prover. 
However, these approaches have inherent limitations in capturing the deep semantic equivalence between informal and formal mathematical statements. 
They may overlook semantic inconsistencies between the two modalities or produce false negatives by incorrectly rejecting semantically valid formalizations \citep{liu2025rethinking}. 
To overcome this, our work adopts Bidirectional Extender Definitional Equivalence (BEq) as the primary evaluation metric \citep{liu2025rethinking}. This choice is motivated by BEq's high alignment with human expert judgment in assessing semantic equivalence. 
The core principle of BEq is to define two formal statements as equivalent if they are mutually transformable through a predefined, finite, and verifiable set of transformation rules \citep{aczel2013homotopy}.

\section{Methodology}

\subsection{Direct Dependency Retrieval}

To prevent hallucination during generation, existing dependency retrieval methods select items from a pre-existing library to form their output \citep{yang2023leandojo,liu2025rethinking}, as illustrated in Figure \ref{fig:example-0} (a). However, this selected-based paradigm suffers from two primary limitations. First, training an encoder to produce effective embeddings for every formal object in the library to maximize retrieval performance is a challenge in itself \citep{xiong2020approximate,gao2023retrieval}. Second, the method is sensitive to superficial variations in the input: semantically equivalent mathematical statements may yield different results in maximum cosine similarity retrieval due to minor differences in their syntactic structure or notational representation \citep{asl2024robustsentembed}. 
Our empirical results in Table \ref{tab:dependency-retrieval-icl} confirm that such embedding-based retrieval methods exhibit low recall and precision. Inaccurate retrieval results don't contribute to the language model in autoformalization tasks. Figure \ref{fig:example-0} (a) illustrates this failure mode: in the absence of a usable dependency from the retrieved Mathlib candidates, the model attempts to construct a local function functionally equivalent to \texttt{factorization}, leading to an erroneous output. Furthermore, this approach lacks scalability, as it cannot effectively leverage the growing volume of available datasets to continuously enhance model capabilities.

\begin{table}
	\centering
	\begin{adjustbox}{width=.5\textwidth,center}
		\begin{tabular}{ccS[table-format=2.3,round-precision=3,round-mode=places]S[table-format=2.3,round-precision=3,round-mode=places]}
			\toprule
			\textbf{Difficulty} &
			\multicolumn{1}{c}{\textbf{Num}} &
			\multicolumn{1}{c}{\textbf{Depend Rate}} &
			\multicolumn{1}{c}{\textbf{Depend Length}}\\
			\midrule
			0  & 835 & 0.18802395209580838 & 0.26706586826347306\\
			1  & 120040 & 0.5550816394535155 & 1.0610963012329224\\
			2  & 113816 & 0.7837562381387503 & 1.8629366697125185\\
			3  & 49841 & 0.80734736461949 & 2.079312212836821\\
			4  & 69882 & 0.7952405483529378 & 1.9577001230645945\\
			5  & 67547 & 0.5233837180037604 & 1.1056893718447895\\
			6  & 41312 & 0.8108781951975214 & 2.2299089852827265\\
			7  & 36675 & 0.7539468302658486 & 2.1081663258350374\\
			8  & 8682 & 0.8195116332642248 & 2.3931121861322278\\
			9  & 708 & 0.8248587570621468 & 2.611581920903955\\
			\bottomrule
		\end{tabular}
	\end{adjustbox}
	\caption{Dataset statistics of FineLeanCorpus. \textbf{Depend Rate} denotes the percentage of samples that have explicit mathlib dependencies within each difficulty level. \textbf{Depend Length} denotes the mean number of explicit mathlib dependency items for samples within each difficulty level. Higher \textbf{Depend Rate} and \textbf{Depend Length} values typically indicate more challenging tasks.}
	\label{tab:data-info-0}
\end{table}

\begin{table*}
	\centering
	\begin{adjustbox}{width=\textwidth,center}
		\begin{tabular}{l|S[table-format=2.2,round-precision=2,round-mode=places]S[table-format=2.2,round-precision=2,round-mode=places]S[table-format=2.2,round-precision=2,round-mode=places]S[table-format=2.2,round-precision=2,round-mode=places]S[table-format=2.2,round-precision=2,round-mode=places]S[table-format=2.2,round-precision=2,round-mode=places]S[table-format=2.2,round-precision=2,round-mode=places]S[table-format=2.2,round-precision=2,round-mode=places]S[table-format=2.2,round-precision=2,round-mode=places]S[table-format=2.2,round-precision=2,round-mode=places]}
			\toprule
			\multirow{2}{*}{\textbf{Method}} &
			\multicolumn{2}{c}{\textbf{\,\, Diff01}} &
			\multicolumn{2}{c}{\textbf{\,\, Diff23}} &
			\multicolumn{2}{c}{\textbf{\,\, Diff45}} &
			\multicolumn{2}{c}{\textbf{\,\, Diff67}} &
			\multicolumn{2}{c}{\textbf{\,\, Diff89}} \\
			\cmidrule(lr){2-3} \cmidrule(lr){4-5} \cmidrule(lr){6-7} \cmidrule(lr){8-9} \cmidrule(lr){10-11}
			& {\bfseries Hall$_\text{m}$}
            & {\bfseries Hall$_\text{std}$}
            & {\bfseries Hall$_\text{m}$}
            & {\bfseries Hall$_\text{std}$}
            & {\bfseries Hall$_\text{m}$}
            & {\bfseries Hall$_\text{std}$}
            & {\bfseries Hall$_\text{m}$}
            & {\bfseries Hall$_\text{std}$}
            & {\bfseries Hall$_\text{m}$}
            & {\bfseries Hall$_\text{std}$} \\
			\midrule

            Claude-3.5-Sonnet
			& 0.3498427672962265
            & 0.3217058968053556
            & 0.300598443223
            & 0.2573053101839686
            & 0.277148809523
            & 0.25870843462684584
            & 0.28479365079350005
            & 0.22076492818566867
            & 0.276160173163
            & 0.22118172274726233
            \\
   
			DeepSeek-R1
			& 0.23458926525773194
            & 0.28236159537551225
            & 0.3301628413569231
            & 0.3133363220603008
            & 0.2653541834656566
            & 0.2891630968814042
            & 0.3065441100865
            & 0.2971994605043282
            & 0.302954725829
            & 0.2885809083243518
            \\
			
			GPT-4o
			& 0.2905303030318182
            & 0.27989000137193565
            & 0.29803418803128207
            & 0.2748659515897591
            & 0.2875448671919598
            & 0.26573672320765096
            & 0.29023689877839204
            & 0.2780079437062113
            & 0.30073389012575763
            & 0.2585760998066564
            \\

            Ling-flash-2.0
			& 0.326612785787156
            & 0.3161762375317554
            & 0.38517565220050004
            & 0.31439379280785773
            & 0.3423849206370001
            & 0.29818853525487327
            & 0.325408019268
            & 0.2767728863854131
            & 0.36206491302900007
            & 0.2766387738545109
            \\
			
			Qwen-Max
			& 0.22980769230673076
            & 0.30286899519033667
            & 0.27435248991326533
            & 0.30576990183206304
            & 0.20936948853333334
            & 0.241521686559319
            & 0.24156388000502513
            & 0.26455260059186486
            & 0.24743612055075379
            & 0.2382686384110975
            \\

            R@5
            & \textbf{\,\,\>0.00}
            & \textbf{\,\,\>0.00}
            & \textbf{\,\,\>0.00}
            & \textbf{\,\,\>0.00}
            & \textbf{\,\,\>0.00}
            & \textbf{\,\,\>0.00}
            & \textbf{\,\,\>0.00}
            & \textbf{\,\,\>0.00}
            & \textbf{\,\,\>0.00}
            & \textbf{\,\,\>0.00}
            \\

            R@10
            & \textbf{\,\,\>0.00}
            & \textbf{\,\,\>0.00}
            & \textbf{\,\,\>0.00}
            & \textbf{\,\,\>0.00}
            & \textbf{\,\,\>0.00}
            & \textbf{\,\,\>0.00}
            & \textbf{\,\,\>0.00}
            & \textbf{\,\,\>0.00}
            & \textbf{\,\,\>0.00}
            & \textbf{\,\,\>0.00}
            \\
			
			DDR
			& 0.016
            & 0.08389182223457682
            & 0.00602725366855346
            & 0.03686358645419893
            & 0.004540420820155038
            & 0.02624359106542632
            & 0.0030626780628205128
            & 0.022527428053560623
            & 0.01373524003109756
            & 0.08741572814878662
			\\
			\bottomrule
		\end{tabular}
	\end{adjustbox}
     \caption{Hallucination rates in retrieved dependency results. {\bfseries Hall$_\text{m}$} measures the mean proportion of hallucinated items per sample in retrieved results; {\bfseries Hall$_\text{std}$} measures the standard deviation. \textbf{R@k} denotes the method from \citet{liu2025rethinking}, where the top-k retrieved results are used as dependency outputs.y}
	\label{tab:dependency-retrieval-hall}
\end{table*}

To address the challenge of LLMs hallucinating non-existent or irrelevant library dependencies during autoformalization, we propose a generation-plus-verification pipeline, with DDR as its core component. DDR is a SFT model to operate in an end-to-end fashion, directly extracting potential library dependencies from an input mathematical statement. The SFT based paradigm enables it to effectively absorb and leverage continuously growing training data, thereby demonstrating excellent scalability. In contrast to conventional two-stage retrieval methods that require additional encoder model \citep{liu2025rethinking}, the design of DDR streamlines the architecture, leading to a simpler, more efficient, and better-performing approach. Accurate dependency retrieval reduces the complexity of the subsequent formalization task, thereby enhancing the overall performance of the autoformalizer. As illustrated in Figure \ref{fig:example-0} (b), providing the model with precise dependencies enables it to more reliably generate the correct formal output, thus increasing the overall success rate.

The effective application of the DDR method entails two primary challenges: first, constructing a high-quality dataset for model fine-tuning, and second, implementing an output filtering mechanism to validate the generated dependencies.
Such challenges can be addressed by the following SAC method.

\subsection{Suffix Array Check}
\label{sec:saa}

To accurately and efficiently verify the existence of dependency objects generated by LLMs within a formal library, we introduce a checking method based on suffix arrays \citep{liu2024infini}. The core idea of this approach is to first concatenate the identifiers of all defined objects in the library, such as pre-proven theorems, into a single long string using a specific delimiter. Subsequently, a suffix array is constructed from this concatenated long string in linear complexity \citep{lee2022deduplicating,karkkainen2006linear}. By definition, a suffix array is an array of integers that stores the starting indices of all suffixes of the original string, sorted in lexicographical order \citep{karkkainen2006linear}. Figure \ref{fig:example-1} illustrates the SAC process with a simplified example, assuming the library contains only two objects and there are three pending items. The sorted nature of the suffix array facilitates efficient matching of these pending items via binary search. Specifically, for each pending item, we can locate the range of all suffixes that have the item as a prefix within the suffix array. This process enables the rapid retrieval of all library objects corresponding to the pending items, thereby completing their existence verification.

SAC is an efficient method for verification, capable of meeting the real-time requirements of large-scale requests \citep{liu2024infini}. For a formal library containing $N$ objects with an average length of $d$, and $M$ pending items with an average length of $s$, conventional string matching methods typically involve brute-force comparison, resulting in a time complexity of $\mathcal{O}\left((s+d)MN\right)$. In contrast, the suffix array auxiliary check operates on a suffix array constructed over the entire library. For each pending item, it utilizes binary search on the suffix array to locate matches. The overall time complexity of this approach is $\mathcal{O}(sM\log N)$. This significant reduction in computational complexity enables real-time responsiveness \citep{liu2024infini}.

\subsection{Constructing High-quality Training Dataset}

Fine-tuning a precise and efficient DDR model necessitates a dataset of pairs, each comprising a mathematical statement and its corresponding library dependencies. However, no such public dataset is currently available, and its construction is primarily challenged by the difficulty of accurately extracting and efficiently verifying library dependencies from existing formal statements. To address this, we employ the above SAC method to process a large-scale autoformalization corpus to generate the required training data. 
We select FineLeanCorpus \citep{peng2025criticlean} as the source corpus, which contains over 500,000 samples spanning a wide range of mathematical domains and difficulty levels. After partitioning the corpus into training and test sets based on problem difficulty, we leverage the efficiency of SAC to process the entire training partition in under two hours. This process yield a high-quality dataset suitable for fine-tuning the DDR model.
This resulting dataset has been made open-access at \url{https://huggingface.co/datasets/Palca/FineLeanCorpus_DDR}. 
We anticipate that this contribution will foster further research on the field of formal mathematical proving.

\section{Direct Dependency Retrieval-augmented Autoformalization}

\begin{table*}
	\centering
	\begin{adjustbox}{width=\textwidth,center}
		\begin{tabular}{l|S[table-format=2.2,round-precision=2,round-mode=places]S[table-format=2.2,round-precision=2,round-mode=places]S[table-format=2.2,round-precision=2,round-mode=places]S[table-format=2.2,round-precision=2,round-mode=places]S[table-format=2.2,round-precision=2,round-mode=places]S[table-format=2.2,round-precision=2,round-mode=places]S[table-format=2.2,round-precision=2,round-mode=places]S[table-format=2.2,round-precision=2,round-mode=places]S[table-format=2.2,round-precision=2,round-mode=places]S[table-format=2.2,round-precision=2,round-mode=places]}
			\toprule
			\multirow{2}{*}{\textbf{Method}} &
			\multicolumn{2}{c}{\textbf{\,\, Diff01}} &
			\multicolumn{2}{c}{\textbf{\,\, Diff23}} &
			\multicolumn{2}{c}{\textbf{\,\, Diff45}} &
			\multicolumn{2}{c}{\textbf{\,\, Diff67}} &
			\multicolumn{2}{c}{\textbf{\,\, Diff89}} \\
			\cmidrule(lr){2-3} \cmidrule(lr){4-5} \cmidrule(lr){6-7} \cmidrule(lr){8-9} \cmidrule(lr){10-11}
			& \textbf{\,\, Precision} & {\textbf{\,\, Recall}} & {\textbf{\,\, Precision}} & {\textbf{\,\, Recall}} &
			{\textbf{\,\, Precision}} & {\textbf{\,\, Recall}} &{\textbf{\,\, Precision}} & {\textbf{\,\, Recall}} &
			{\textbf{\,\, Precision}} & {\textbf{\,\, Recall}} \\
			\midrule
			
			Claude-3.5-Sonnet
			& 0.49024999999999996
			& 0.5235
			& 0.27716071428571426
			& 0.38476190476190475
			& 0.214
			& 0.3403333333333333
			& 0.2798333333333333
			& 0.4347103174603174
			& 0.27285335497835495
			& 0.3339900793650794
			\\
			
			DeepSeek-R1
			& 0.5318333333333333
			& 0.5578333333333333
			& 0.240378663003663
			& 0.35368452380952375
			& 0.19317460317460317
			& 0.323297619047619
			& 0.23088737161531278
			& 0.36863888888888885
			& 0.2622460317460317
			& 0.3715277777777778
			\\
			
			GPT-4o
			& 0.5387500000000001
			& 0.5480833333333334
			& 0.2828809523809524
			& 0.3174702380952381
			& 0.20350000000000001
			& 0.24749999999999997
			& 0.2624166666666667
			& 0.3186984126984127
			& 0.2743055555555555
			& 0.28998015873015875
			\\
			
			Ling-flash-2.0
			& 0.44925
			& 0.4693333333333334
			& 0.1738511904761905
			& 0.23617857142857143
			& 0.14754166666666665
			& 0.21949999999999995
			& 0.21478012265512267
			& 0.3199484126984127
			& 0.181663295038295
			& 0.24742658730158731
			\\
			
			Qwen-Max
			& 0.4794642857142857
			& 0.51125
			& 0.26315873015873015
			& 0.2933809523809524
			& 0.2278333333333333
			& 0.27658333333333335
			& 0.26225
			& 0.3313809523809524
			& 0.2663928571428571
			& 0.32313293650793645
			\\

           R@5
			& 0.022000000000000002
			& 0.06916666666666665
			& 0.073
			& 0.15097023809523807
			& 0.048999999999999995
			& 0.11925000000000001
			& 0.079
			& 0.17116666666666666
			& 0.085
			& 0.14372222222222222
			\\
			
			R@10
			& 0.016500000000000004
			& 0.10041666666666665
			& 0.04449999999999999
			& 0.19213690476190476
			& 0.032
			& 0.15058333333333332
			& 0.059500000000000004
			& 0.26547619047619053
			& 0.061
			& 0.2064801587301587
			\\
			DDR
			& \textbf{\,\,\>0.84}
			& \textbf{\,\,\>0.82}
			& \textbf{\,\,\>0.90}
			& \textbf{\,\,\>0.87}
			& \textbf{\,\,\>0.91}
			& \textbf{\,\,\>0.92}
			& \textbf{\,\,\>0.91}
			& \textbf{\,\,\>0.88}
			& \textbf{\,\,\>0.83}
			& \textbf{\,\,\>0.83}
			\\
			\bottomrule
		\end{tabular}
	\end{adjustbox}
     \caption{Comparison of the retrieval dependency results. Results shown are filtered (i.e., without hallucinations).}
	\label{tab:dependency-retrieval-icl}
\end{table*}

\begin{figure*}
	\centering
	\includegraphics[width=1.\textwidth]{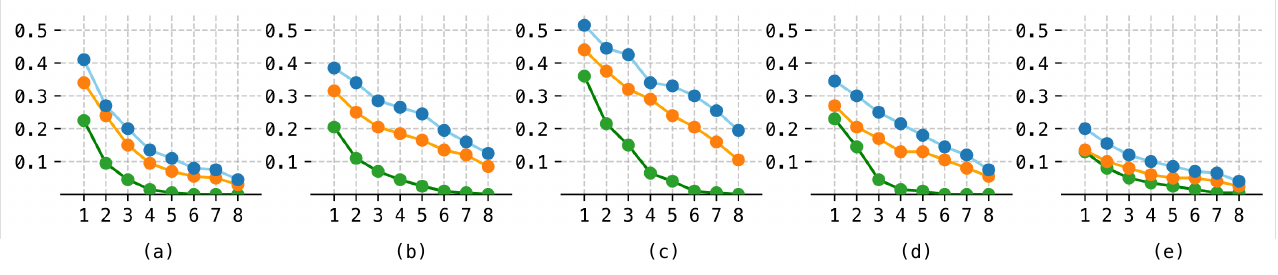}
	\caption{LLM consistency across multiple attempts. Subplots (a–e) correspond to the performance of DeepSeek-R1 on datasets Diff01, Diff23, Diff45, Diff67, and Diff89, respectively. Each subplot illustrates the proportion of problems ($y$ axis) for which at least k (from 1 to 8, $x$ axis) out of 8 attempts successfully pass the BEq verification.
		Dependency retrieval method:
		\textcolor{blue}{DDR (blue)}; \textcolor{orange}{ICL (orange)}; \textcolor{lightgreen}{N/A (green)}.
	}
	\label{fig:example-deepseek_r1_main2}
\end{figure*}

This section presents a comprehensive evaluation of the proposed method. Our evaluation is two-fold: first, we directly assess the performance of DDR on the dependency retrieval task. Second, we measure the improvements it brings by evaluating the pass@8 success rate on the downstream autoformalization task. 
Furthermore, the experiments provide additional insights, such as a discussion on hallucinations in ICL methods and an analysis of the difficulty of autoformalization datasets.

\subsection{Dataset Overview and Difficulty Analysis}

The {difficulty} attribute is a key metric in FineLeanCorpus designed to quantify the challenge of formalization, rated on an 11-point scale from 0 to 10. 
In our experimental setup, we consolidated samples with a {difficulty} of 10 into the {difficulty} 9 category.
The statistical results of Mathlib dependency on datasets are summarized in Table \ref{tab:data-info-0}.

We randomly sampled 100 statements from each difficulty category and then combined the samples from every two adjacent difficulty categories as the test set. 
This process resulted in five test sets, named \textbf{Diff01}, \textbf{Diff23}, \textbf{Diff45}, \textbf{Diff67}, and \textbf{Diff89}, where \textbf{Diff\textit{ij}} denotes the combined set of samples from difficulty levels \textbf{\textit{i}} and \textbf{\textit{j}}, with each test set containing 200 statements. All remaining samples in the dataset were used to fine-tune the DDR model.

The original difficulty ratings for the FineLeanCorpus \citep{peng2025criticlean} datasets, assigned by an LLM grader, exhibit a notable discrepancy with our experimental results in Table \ref{tab:results-3} and Table \ref{tab:results-icl}. 
To investigate this inconsistency, we propose assessing the actual formalization difficulty using two alternative metrics: \textbf{Depend Rate} and \textbf{Depend Length}. 
Based on the data in Table \ref{tab:data-info-0}, we can calculate the weighted averages of \textbf{Depend Rate} and \textbf{Depend Length} for the \textbf{Diff23}, \textbf{Diff45}, \textbf{Diff67}, and \textbf{Diff89} subsets. 
The results are as follows: \textbf{Diff45} has the lowest values (0.662, 1.539), the values for \textbf{Diff23} (0.791, 1.929) and \textbf{Diff67} (0.784, 2.173) are higher, and \textbf{Diff89} has the highest values (0.820, 2.410).
These metrics reveal a difficulty ranking of \textbf{Diff89} > \textbf{Diff67} > \textbf{Diff23} > \textbf{Diff45}. This ranking aligns closely with our experimental observations, suggesting that \textbf{Depend Rate} and \textbf{Depend Length}, derived via our DDR and SAC, can serve as a potential indicator of formalization difficulty along with LLM-based graders. 
The \textbf{Diff01} subset is excluded from this analysis because its imprecise informal descriptions introduce extraneous formalization challenges, making its difficulty hard to assess with these metrics. Detailed discussion are presented in appendix \ref{append:examples-illustration}

\begin{figure*}
	\centering
	\includegraphics[width=1.\textwidth]{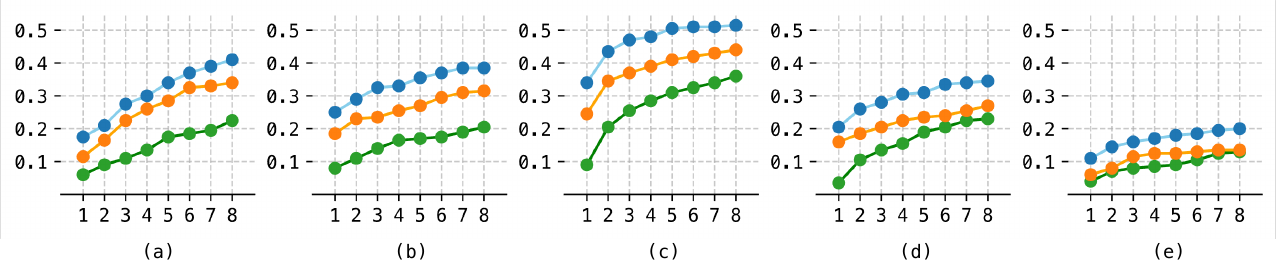}
	\caption{LLM efficiency across multiple attempts. Subplots (a–e) correspond to the performance of DeepSeek-R1 on datasets Diff01, Diff23, Diff45, Diff67, and Diff89, respectively. Each subplot illustrates the intermediate pass@k score ($y$ axis) out of 8 attempts using BEq verification, k range from 1 to 8 ($x$ axis).
     Dependency retrieval method:
		\textcolor{blue}{DDR (blue)}; \textcolor{orange}{ICL (orange)}; \textcolor{lightgreen}{N/A (green)}.
	}
	\label{fig:example-deepseek_r1_main3}
\end{figure*}

\begin{table*}
	\centering
	\begin{adjustbox}{width=\textwidth,center}
		\begin{tabular}{l|S[table-format=2.3,round-precision=3,round-mode=places]S[table-format=2.3,round-precision=3,round-mode=places]S[table-format=2.3,round-precision=3,round-mode=places]S[table-format=2.3,round-precision=3,round-mode=places]S[table-format=2.3,round-precision=3,round-mode=places]S[table-format=2.3,round-precision=3,round-mode=places]S[table-format=2.3,round-precision=3,round-mode=places]S[table-format=2.3,round-precision=3,round-mode=places]S[table-format=2.3,round-precision=3,round-mode=places]S[table-format=2.3,round-precision=3,round-mode=places]}
			\toprule
			\multirow{2}{*}{\textbf{Method}} 
            &
			\multicolumn{2}{c}{\textbf{\,\, Diff01}} &
			\multicolumn{2}{c}{\textbf{\,\, Diff23}} &
			\multicolumn{2}{c}{\textbf{\,\, Diff45}} &
			\multicolumn{2}{c}{\textbf{\,\, Diff67}} &
			\multicolumn{2}{c}{\textbf{\,\, Diff89}} \\
			\cmidrule(lr){2-3} \cmidrule(lr){4-5} \cmidrule(lr){6-7} \cmidrule(lr){8-9} \cmidrule(lr){10-11}
			& \textbf{\,\, TypeC} & {\textbf{\,\, BEq}} & {\textbf{\,\, TypeC}} & {\textbf{\,\, BEq}} &
			{\textbf{\,\, TypeC}} & {\textbf{\,\, BEq}} &{\textbf{\,\, TypeC}} & {\textbf{\,\, BEq}} &
			{\textbf{\,\, TypeC}} & {\textbf{\,\, BEq}} \\
			\midrule
			
			MMA
& 0.47
& 0.095
& 0.455
& 0.11
& 0.6
& 0.19
& 0.5
& 0.1
& 0.385
& 0.065
			\\
			
			RA
& 0.67
& 0.115
& 0.730
& 0.170
& 0.765
& 0.295
& 0.735
& 0.185
& 0.61
& 0.105
			\\
			
			RA$\texttt{+}$R@5
& 0.86
& 0.185
& 0.795
& 0.18
& 0.805
& 0.145
& 0.825
& 0.115
& 0.765
& 0.135
			\\
			
			RA$\texttt{+}$R@10
& \textbf{\,\,\>0.880}
& 0.205
& 0.86
& 0.2
& \textbf{\,\,\>0.910}
& 0.255
& 0.825
& 0.16
& 0.765
& \textbf{\,\,\>0.150}
			\\
			
			RA$\texttt{+}$DDR
& 0.865
& \textbf{\,\,\>0.235}
& \textbf{\,\,\>0.900}
& \textbf{\,\,\>0.215}
& \textbf{\,\,\>0.910}
& \textbf{\,\,\>0.305}
& \textbf{\,\,\>0.865}
& \textbf{\,\,\>0.225}
& \textbf{\,\,\>0.770}
& \textbf{\,\,\>0.150}

			\\
   
			\bottomrule
		\end{tabular}
	\end{adjustbox}
     \caption{Comparison of the autoformalization results using SFT methods.
		\textbf{MMA} denotes the LLM fine-tuned on MMA’s Lean subset. 
		\textbf{RA} refers to the pure autoformalization method from \citet{liu2025rethinking}, while \textbf{R@5} and \textbf{R@10} denote its retrieval-augmented variants using the top-5 and top-10 retrieved results, respectively.
		\textbf{RA+DDR} denotes the new retrieval-augmented variant incorporating the proposed DDR.}
	\label{tab:results-3}
\end{table*}

\subsection{Setup}

Due to space constraints, we defer the detailed description of our experimental setup to the appendix \ref{append:experiment-details}. 
Our evaluation utilizes two primary metrics: Type Checking (TypeC) and Bidirectional Extended Definitional Equality (BEq). We compare our method against a comprehensive set of baseline methods, including MMA \citep{jiang2023multilingual}, DeepSeek-R1 \citep{guo2025deepseek}, Ling-flash-2.0 \citep{ai2025ming}, Claude-3.5-Sonnet \citep{anthropic_claude35_sonnet_2024}, GPT-4o \citep{achiam2023gpt}, Qwen-Max \citep{qwen2025qwen25technicalreport}, as well as RA+R@5 and RA+R@10 \citep{liu2025rethinking}.

\subsection{Hallucination Study}

Prior to the introduction of DDR, a sound mechanism for assessing dependency hallucinations in the outputs of LLMs on autoformalization tasks was lacking. The proposed DDR framework, through the integrated SAC, provides a practical and efficient solution to this problem. This method determines whether each generated dependency exists within the formal library, thereby enabling the quantification of hallucination severity. Define the hallucination rate (\textbf{Hall}) as the proportion of hallucinatory items in a given set of dependencies. For instance, if a model outputs (\texttt{Real.div, Real.max, Nat.succ}) and only \texttt{Nat.succ} is authentic, the \textbf{Hall} is 2/3. $\textbf{Hall}_\textbf{m}$ and $\textbf{Hall}_\textbf{std}$ denote the mean and standard deviation of this rate across the test set, respectively. The hallucination rates for various mainstream dependency retrieval methods are presented in Table \ref{tab:dependency-retrieval-hall}.

The analysis reveals that R@5 and R@10, as select-based methods, are inherently free from hallucination by design, since their outputs are drawn directly from the formal library. In contrast, all generative retrieval methods are susceptible to hallucination. Our proposed DDR method exhibits outstanding performance, with a mean hallucination rate ($\textbf{Hall}_\textbf{m}$) below 0.02 and a standard deviation ($\textbf{Hall}_\textbf{std}$) below 0.09, indicating nearly non-existent hallucination. In stark contrast, other ICL-based methods suffer from severe hallucination, with both the mean and standard deviation of their \textbf{Hall} values hovering around 0.30. This implies that, on average, approximately 30\% of the dependencies generated by these methods are hallucinatory. 

\begin{table*}
	\centering
	\begin{adjustbox}{width=\textwidth,center}
		\begin{tabular}{l|lS[table-format=2.3,round-precision=3,round-mode=places]S[table-format=2.3,round-precision=3,round-mode=places]S[table-format=2.3,round-precision=3,round-mode=places]S[table-format=2.3,round-precision=3,round-mode=places]S[table-format=2.3,round-precision=3,round-mode=places]S[table-format=2.3,round-precision=3,round-mode=places]S[table-format=2.3,round-precision=3,round-mode=places]S[table-format=2.3,round-precision=3,round-mode=places]S[table-format=2.3,round-precision=3,round-mode=places]S[table-format=2.3,round-precision=3,round-mode=places]}
			\toprule
			\multirow{2}{*}{\textbf{Method}} & \multirow{2}{*}{\textbf{RAG}} &
			\multicolumn{2}{c}{\textbf{\,\, Diff01}} &
			\multicolumn{2}{c}{\textbf{\,\, Diff23}} &
			\multicolumn{2}{c}{\textbf{\,\, Diff45}} &
			\multicolumn{2}{c}{\textbf{\,\, Diff67}} &
			\multicolumn{2}{c}{\textbf{\,\, Diff89}} \\
			\cmidrule(lr){3-4} \cmidrule(lr){5-6} \cmidrule(lr){7-8} \cmidrule(lr){9-10} \cmidrule(lr){11-12}
			& & \textbf{\,\, TypeC} & {\textbf{\,\, BEq}} & {\textbf{\,\, TypeC}} & {\textbf{\,\, BEq}} &
			{\textbf{\,\, TypeC}} & {\textbf{\,\, BEq}} &{\textbf{\,\, TypeC}} & {\textbf{\,\, BEq}} &
			{\textbf{\,\, TypeC}} & {\textbf{\,\, BEq}} \\
			\midrule
   
			Claude-3.5-Sonnet
            & N/A
& 0.83
& 0.415
& 0.74
& 0.33
& \textbf{\,\,\>0.775}
& 0.385
& \textbf{\,\,\>0.680}
& 0.205
& 0.53
& 0.1
			\\

			Claude-3.5-Sonnet
            & ICL
& 0.74
& 0.37
& 0.515
& 0.245
& 0.64
& 0.33
& 0.455
& 0.145
& 0.385
& 0.11
			\\

           Claude-3.5-Sonnet
            & DDR
& \textbf{\,\,\>0.905}
& \textbf{\,\,\>0.460}
& \textbf{\,\,\>0.770}
& \textbf{\,\,\>0.335}
& 0.765
& \textbf{\,\,\>0.425}
& 0.625
& \textbf{\,\,\>0.220}
& \textbf{\,\,\>0.630}
& \textbf{\,\,\>0.130}
			\\ [2pt]
			
			\hdashline \\ [-10pt]
			DeepSeek-R1
            & N/A
& 0.91
& 0.225
& 0.745
& 0.205
& 0.82
& 0.36
& 0.78
& 0.23
& 0.66
& 0.13
			\\

			DeepSeek-R1
            & ICL
& 0.88
& 0.34
& 0.755
& 0.315
& 0.825
& 0.44
& 0.76
& 0.27
& 0.57
& 0.135
			\\

			DeepSeek-R1
            & DDR
& \textbf{\,\,\>0.940}
& \textbf{\,\,\>0.410}
& \textbf{\,\,\>0.875}
& \textbf{\,\,\>0.385}
& \textbf{\,\,\>0.940}
& \textbf{\,\,\>0.515}
& \textbf{\,\,\>0.875}
& \textbf{\,\,\>0.345}
& \textbf{\,\,\>0.695}
& \textbf{\,\,\>0.200}
			\\ [2pt]
			
			\hdashline \\ [-10pt]
			
			GPT-4o
            & N/A
& 0.905
& 0.44
& 0.76
& 0.295
& 0.8
& 0.4
& 0.745
& 0.24
& 0.575
& 0.135
			\\

			GPT-4o
            & ICL
& 0.88
& \textbf{\,\,\>0.460}
& 0.675
& 0.255
& 0.73
& 0.345
& 0.625
& 0.21
& 0.405
& 0.12
			\\

			GPT-4o
            & DDR
& \textbf{\,\,\>0.940}
& \textbf{\,\,\>0.460}
& \textbf{\,\,\>0.775}
& \textbf{\,\,\>0.320}
& \textbf{\,\,\>0.850}
& \textbf{\,\,\>0.455}
& \textbf{\,\,\>0.780}
& \textbf{\,\,\>0.275}
& \textbf{\,\,\>0.630}
& \textbf{\,\,\>0.150}
			\\ [2pt]
			
			\hdashline \\ [-10pt]
			
			Ling-flash-2.0
            & N/A
& \textbf{\,\,\>0.850}
& \textbf{\,\,\>0.405}
& \textbf{\,\,\>0.610}
& 0.21
& \textbf{\,\,\>0.665}
& 0.3
& 0.51
& 0.19
& 0.38
& 0.085
			\\

			Ling-flash-2.0
            & ICL
& 0.775
& 0.355
& 0.42
& 0.17
& 0.59
& 0.28
& 0.395
& 0.17
& 0.295
& 0.085
			\\

			Ling-flash-2.0
            & DDR
& 0.775
& 0.4
& 0.595
& \textbf{\,\,\>0.225}
& 0.63
& \textbf{\,\,\>0.360}
& \textbf{\,\,\>0.535}
& \textbf{\,\,\>0.205}
& \textbf{\,\,\>0.440}
& \textbf{\,\,\>0.115}
			\\ [2pt]
			
			\hdashline \\ [-10pt]
			Qwen-Max
            & N/A
& 0.795
& \textbf{\,\,\>0.375}
& 0.605
& 0.22
& \textbf{\,\,\>0.665}
& \textbf{\,\,\>0.360}
& 0.62
& 0.205
& 0.44
& 0.105
			\\ 
			
			Qwen-Max
            & ICL
& 0.73
& 0.315
& 0.445
& 0.16
& 0.55
& 0.26
& 0.445
& 0.165
& 0.35
& 0.115
			\\ 	
			Qwen-Max
            & DDR
& \textbf{\,\,\>0.830}
& \textbf{\,\,\>0.375}
& \textbf{\,\,\>0.660}
& \textbf{\,\,\>0.250}
& 0.56
& 0.285
& \textbf{\,\,\>0.695}
& \textbf{\,\,\>0.255}
& \textbf{\,\,\>0.545}
& \textbf{\,\,\>0.120}
			\\
   
			\bottomrule
		\end{tabular}
	\end{adjustbox}
     \caption{Comparison of the autoformalization results using ICL methods. \textbf{RAG} column indicates the dependency retrieval strategy used in autoformalization: N/A denotes no retrieval, ICL denotes prompt-based retrieval, and DDR refers to the proposed method.}
	\label{tab:results-icl}
\end{table*}

\subsection{Retrieval Results}
Although R@5 and R@10 are hallucination-free, their retrieval performance is questionable. To isolate retrieval quality from the issue of hallucination and enable a fair comparison of the methods' true capabilities, we first filtered out all hallucinatory items from each method's output and then recalculated their precision and recall. The performance comparison after filtering is shown in Table \ref{tab:dependency-retrieval-icl}.

After filtering, the results show substantial performance disparities among the models. Despite having no hallucinations, R@5 and R@10 yield the lowest precision (0.02-0.10) and recall (0.10-0.20), indicating their retrieval capability is severely limited. The filtered prompt-based methods achieve moderate performance, with precision scores ranging from 0.2 to 0.5 and recall scores from 0.3 to 0.5. Our proposed DDR method demonstrates the most superior performance, achieving a high precision of 0.8-0.9 and a recall of 0.8-0.9, significantly outperforming all other baseline methods.

\subsection{Autoformalization Results}

The efficacy of DDR is further substantiated by its performance improvements in the downstream autoformalization task. As shown in Table \ref{tab:results-3} and Table \ref{tab:results-icl}, integrating DDR yields significant performance gains under both SFT and ICL frameworks. We specifically evaluated the combination of DDR with the RA model, an advanced model fine-tuned for autoformalization in \citet{liu2025rethinking}. A format mismatch exists, as the RA model requires a comprehensive input format (full dependency name, informal description, and code), whereas DDR outputs only abbreviated dependency names. Consequently, the R@5 and R@10 retrievers from \citet{liu2025rethinking}, designed for RA, are more format-compatible. Despite this mismatch, the combination of RA and DDR consistently outperforms all baseline methods. This success is primarily attributed to DDR's superior retrieval precision and recall, highlighting its excellent generalization and robustness. In contrast, the performance of R@5 and R@10 reflects a trade-off between precision and recall, consistent with their respective retrieval characteristics.

To evaluate DDR's potential as a plug-and-play component, we integrated it with ICL methods. Table \ref{tab:results-icl} compares three experimental setups: (1) direct autoformalization, (2) two-stage autoformalization using a prompt-based dependency retrieval, and (3) two-stage autoformalization using a DDR-based dependency retrieval. The results indicate that the DDR-based RAG is the superior strategy in most test scenarios. Experiments on Deepseek-R1, illustrated in Figure \ref{fig:example-deepseek_r1_main2}, further reveal DDR's stability advantage: the DDR-integrated method not only achieves a higher total pass@8 success count but also exhibits a significantly better distribution of successful attempts across 8 independent trials compared to the baseline. Furthermore, Figure \ref{fig:example-deepseek_r1_main3} shows that DDR can complete more tasks with fewer attempts. 


\section{Conclusion}

We propose DDR, a method that leverages the contextual understanding of LLMs to directly generate candidate dependencies for a given informal statement in autoformalization tasks. 
This is complemented by SAC, which utilize suffix array for efficient dependency verification.
Experimental results validate the effectiveness of our approach and demonstrate its potential in ICL hallucinations analysis and in assessing the difficulty of autoformalization datasets.


\section{Limitations}

A key limitation of DDR is its inability to precisely locate the generated items within the library. Specifically, whereas the full retrieval results in methods like \citet{liu2025rethinking} typically include complete metadata such as the item's name, code usage, and documentation, DDR's potential to generate abbreviated or shortcut names precludes the retrieval of this supplementary information.
This discrepancy renders DDR not directly compatible with the retrieval-augmented autoformalizer framework in \citet{liu2025rethinking}. 

DDR also exhibits limitations in handling certain special cases, which necessitate dedicated mechanisms. For instance, some core constructs (such as \texttt{MOD}), despite being crucial and frequently used in formalization, may not exist as standard library items

Formal mathematics libraries are continuously evolving, with dependency items being added, removed, or renamed across different versions. Such changes require corresponding maintenance of our system to ensure its continued functionality and effectiveness.

\bibliography{icml2026_conference}

@article{team2024qwen2,
	title={Qwen2 technical report},
	author={Team, Qwen and others},
	journal={arXiv preprint arXiv:2407.10671},
	volume={2},
	number={3},
	year={2024}
}

@article{hurst2024gpt,
	title={Gpt-4o system card},
	author={Hurst, Aaron and Lerer, Adam and Goucher, Adam P and Perelman, Adam and Ramesh, Aditya and Clark, Aidan and Ostrow, AJ and Welihinda, Akila and Hayes, Alan and Radford, Alec and others},
	journal={arXiv preprint arXiv:2410.21276},
	year={2024}
}

@inproceedings{moura2021lean,
  title={The lean 4 theorem prover and programming language},
  author={Moura, Leonardo de and Ullrich, Sebastian},
  booktitle={International Conference on Automated Deduction},
  pages={625--635},
  year={2021},
  organization={Springer}
}

@book{bertot2013interactive,
  title={Interactive theorem proving and program development: Coq’Art: the calculus of inductive constructions},
  author={Bertot, Yves and Cast{\'e}ran, Pierre},
  year={2013},
  publisher={Springer Science \& Business Media}
}

@book{nipkow2002isabelle,
  title={Isabelle/HOL: a proof assistant for higher-order logic},
  author={Nipkow, Tobias and Wenzel, Markus and Paulson, Lawrence C},
  year={2002},
  publisher={Springer}
}

@article{nawaz2019survey,
  title={A survey on theorem provers in formal methods},
  author={Nawaz, M Saqib and Malik, Moin and Li, Yi and Sun, Meng and Lali, M},
  journal={arXiv preprint arXiv:1912.03028},
  year={2019}
}

@article{li2024survey,
  title={A survey on deep learning for theorem proving},
  author={Li, Zhaoyu and Sun, Jialiang and Murphy, Logan and Su, Qidong and Li, Zenan and Zhang, Xian and Yang, Kaiyu and Si, Xujie},
  journal={arXiv preprint arXiv:2404.09939},
  year={2024}
}

@inproceedings{liu2025rethinking,
  title={Rethinking and improving autoformalization: towards a faithful metric and a dependency retrieval-based approach},
  author={Liu, Qi and Zheng, Xinhao and Lu, Xudong and Cao, Qinxiang and Yan, Junchi},
  booktitle={The Thirteenth International Conference on Learning Representations},
  year={2025}
}

@article{lin2025goedel,
  title={Goedel-prover-v2: Scaling formal theorem proving with scaffolded data synthesis and self-correction},
  author={Lin, Yong and Tang, Shange and Lyu, Bohan and Yang, Ziran and Chung, Jui-Hui and Zhao, Haoyu and Jiang, Lai and Geng, Yihan and Ge, Jiawei and Sun, Jingruo and others},
  journal={arXiv preprint arXiv:2508.03613},
  year={2025}
}

@article{lin2024fvel,
  title={FVEL: Interactive formal verification environment with large language models via theorem proving},
  author={Lin, Xiaohan and Cao, Qingxing and Huang, Yinya and Wang, Haiming and Lu, Jianqiao and Liu, Zhengying and Song, Linqi and Liang, Xiaodan},
  journal={Advances in Neural Information Processing Systems},
  volume={37},
  pages={54932--54946},
  year={2024}
}

@article{zhang2025deeptheorem,
  title={Deeptheorem: Advancing llm reasoning for theorem proving through natural language and reinforcement learning},
  author={Zhang, Ziyin and Xu, Jiahao and He, Zhiwei and Liang, Tian and Liu, Qiuzhi and Li, Yansi and Song, Linfeng and Liang, Zhenwen and Zhang, Zhuosheng and Wang, Rui and others},
  journal={arXiv preprint arXiv:2505.23754},
  year={2025}
}

@article{zhou2024don,
  title={Don't Trust: Verify--Grounding LLM Quantitative Reasoning with Autoformalization},
  author={Zhou, Jin Peng and Staats, Charles and Li, Wenda and Szegedy, Christian and Weinberger, Kilian Q and Wu, Yuhuai},
  journal={arXiv preprint arXiv:2403.18120},
  year={2024}
}

@article{cao2025informal,
  title={From Informal to Formal--Incorporating and Evaluating LLMs on Natural Language Requirements to Verifiable Formal Proofs},
  author={Cao, Jialun and Lu, Yaojie and Li, Meiziniu and Ma, Haoyang and Li, Haokun and He, Mengda and Wen, Cheng and Sun, Le and Zhang, Hongyu and Qin, Shengchao and others},
  journal={arXiv preprint arXiv:2501.16207},
  year={2025}
}

@article{peng2025criticlean,
  title={Criticlean: Critic-guided reinforcement learning for mathematical formalization},
  author={Peng, Zhongyuan and Yao, Yifan and Ma, Kaijing and Guo, Shuyue and Li, Yizhe and Zhang, Yichi and Zhang, Chenchen and Zhang, Yifan and Yu, Zhouliang and Li, Luming and others},
  journal={arXiv preprint arXiv:2507.06181},
  year={2025}
}

@article{yu2025formalmath,
  title={Formalmath: Benchmarking formal mathematical reasoning of large language models},
  author={Yu, Zhouliang and Peng, Ruotian and Ding, Keyi and Li, Yizhe and Peng, Zhongyuan and Liu, Minghao and Zhang, Yifan and Yuan, Zheng and Xin, Huajian and Huang, Wenhao and others},
  journal={arXiv preprint arXiv:2505.02735},
  year={2025}
}

@inproceedings{miranda2025veribench,
  title={Veribench: End-to-end formal verification benchmark for ai code generation in lean 4},
  author={Miranda, Brando and Zhou, Zhanke and Nie, Allen and Obbad, Elyas and Aniva, Leni and Fronsdal, Kai and Kirk, Weston and Soylu, Dilara and Yu, Andrea and Li, Ying and others},
  booktitle={2nd AI for Math Workshop@ ICML 2025},
  year={2025}
}

@article{azerbayev2023proofnet,
  title={Proofnet: Autoformalizing and formally proving undergraduate-level mathematics},
  author={Azerbayev, Zhangir and Piotrowski, Bartosz and Schoelkopf, Hailey and Ayers, Edward W and Radev, Dragomir and Avigad, Jeremy},
  journal={arXiv preprint arXiv:2302.12433},
  year={2023}
}

@article{zheng2021minif2f,
  title={Minif2f: a cross-system benchmark for formal olympiad-level mathematics},
  author={Zheng, Kunhao and Han, Jesse Michael and Polu, Stanislas},
  journal={arXiv preprint arXiv:2109.00110},
  year={2021}
}

@article{ying2024lean,
  title={Lean workbook: A large-scale lean problem set formalized from natural language math problems},
  author={Ying, Huaiyuan and Wu, Zijian and Geng, Yihan and Wang, Jiayu and Lin, Dahua and Chen, Kai},
  journal={Advances in Neural Information Processing Systems},
  volume={37},
  pages={105848--105863},
  year={2024}
}

@article{kimina_prover_2025,
    title = {Kimina-Prover Preview: Towards Large Formal Reasoning Models with Reinforcement Learning},
    author = {Wang, Haiming and Unsal, Mert and Lin, Xiaohan and Baksys, Mantas and Liu, Junqi and Santos, Marco Dos and Sung, Flood and Vinyes, Marina and Ying, Zhenzhe and Zhu, Zekai and Lu, Jianqiao and Saxcé, Hugues de and Bailey, Bolton and Song, Chendong and Xiao, Chenjun and Zhang, Dehao and Zhang, Ebony and Pu, Frederick and Zhu, Han and Liu, Jiawei and Bayer, Jonas and Michel, Julien and Yu, Longhui and Dreyfus-Schmidt, Léo and Tunstall, Lewis and Pagani, Luigi and Machado, Moreira and Bourigault, Pauline and Wang, Ran and Polu, Stanislas and Barroyer, Thibaut and Li, Wen-Ding and Niu, Yazhe and Fleureau, Yann and Hu, Yangyang and Yu, Zhouliang and Wang, Zihan and Yang, Zhilin and Liu, Zhengying and Li, Jia},
    year = {2025},
    url = {http://arxiv.org/abs/2504.11354},
}

@article{yang2023leandojo,
  title={Leandojo: Theorem proving with retrieval-augmented language models},
  author={Yang, Kaiyu and Swope, Aidan and Gu, Alex and Chalamala, Rahul and Song, Peiyang and Yu, Shixing and Godil, Saad and Prenger, Ryan J and Anandkumar, Animashree},
  journal={Advances in Neural Information Processing Systems},
  volume={36},
  pages={21573--21612},
  year={2023}
}

@article{wang2025aria,
  title={Aria: An Agent For Retrieval and Iterative Auto-Formalization via Dependency Graph},
  author={Wang, Hanyu and Xie, Ruohan and Wang, Yutong and Gao, Guoxiong and Yu, Xintao and Dong, Bin},
  journal={arXiv preprint arXiv:2510.04520},
  year={2025}
}

@article{wu2022autoformalization,
  title={Autoformalization with large language models},
  author={Wu, Yuhuai and Jiang, Albert Qiaochu and Li, Wenda and Rabe, Markus and Staats, Charles and Jamnik, Mateja and Szegedy, Christian},
  journal={Advances in neural information processing systems},
  volume={35},
  pages={32353--32368},
  year={2022}
}

@article{zhang2024consistent,
  title={Consistent autoformalization for constructing mathematical libraries},
  author={Zhang, Lan and Quan, Xin and Freitas, Andre},
  journal={arXiv preprint arXiv:2410.04194},
  year={2024}
}

@article{chen2025seed,
  title={Seed-prover: Deep and broad reasoning for automated theorem proving},
  author={Chen, Luoxin and Gu, Jinming and Huang, Liankai and Huang, Wenhao and Jiang, Zhicheng and Jie, Allan and Jin, Xiaoran and Jin, Xing and Li, Chenggang and Ma, Kaijing and others},
  journal={arXiv preprint arXiv:2507.23726},
  year={2025}
}

@article{xin2024deepseek,
  title={Deepseek-prover: Advancing theorem proving in llms through large-scale synthetic data},
  author={Xin, Huajian and Guo, Daya and Shao, Zhihong and Ren, Zhizhou and Zhu, Qihao and Liu, Bo and Ruan, Chong and Li, Wenda and Liang, Xiaodan},
  journal={arXiv preprint arXiv:2405.14333},
  year={2024}
}

@article{xin2024deepseekv15,
  title={Deepseek-prover-v1. 5: Harnessing proof assistant feedback for reinforcement learning and monte-carlo tree search},
  author={Xin, Huajian and Ren, ZZ and Song, Junxiao and Shao, Zhihong and Zhao, Wanjia and Wang, Haocheng and Liu, Bo and Zhang, Liyue and Lu, Xuan and Du, Qiushi and others},
  journal={arXiv preprint arXiv:2408.08152},
  year={2024}
}

@article{ren2025deepseek,
  title={Deepseek-prover-v2: Advancing formal mathematical reasoning via reinforcement learning for subgoal decomposition},
  author={Ren, ZZ and Shao, Zhihong and Song, Junxiao and Xin, Huajian and Wang, Haocheng and Zhao, Wanjia and Zhang, Liyue and Fu, Zhe and Zhu, Qihao and Yang, Dejian and others},
  journal={arXiv preprint arXiv:2504.21801},
  year={2025}
}

@article{liu2025proofaug,
  title={ProofAug: Efficient Neural Theorem Proving via Fine-grained Proof Structure Analysis},
  author={Liu, Haoxiong and Sun, Jiacheng and Li, Zhenguo and Yao, Andrew C},
  journal={arXiv preprint arXiv:2501.18310},
  year={2025}
}

@article{murphy2024autoformalizing,
  title={Autoformalizing euclidean geometry},
  author={Murphy, Logan and Yang, Kaiyu and Sun, Jialiang and Li, Zhaoyu and Anandkumar, Anima and Si, Xujie},
  journal={arXiv preprint arXiv:2405.17216},
  year={2024}
}

@article{poiroux2024improving,
  title={Improving autoformalization using type checking},
  author={Poiroux, Auguste and Weiss, Gail and Kun{\v{c}}ak, Viktor and Bosselut, Antoine},
  journal={arXiv preprint arXiv:2406.07222},
  year={2024}
}

@inproceedings{wang2020exploration,
  title={Exploration of neural machine translation in autoformalization of mathematics in Mizar},
  author={Wang, Qingxiang and Brown, Chad and Kaliszyk, Cezary and Urban, Josef},
  booktitle={Proceedings of the 9th ACM SIGPLAN International Conference on Certified Programs and Proofs},
  pages={85--98},
  year={2020}
}

@article{mahdavi2025leveraging,
  title={Leveraging online olympiad-level math problems for llms training and contamination-resistant evaluation},
  author={Mahdavi, Sadegh and Li, Muchen and Liu, Kaiwen and Thrampoulidis, Christos and Sigal, Leonid and Liao, Renjie},
  journal={arXiv preprint arXiv:2501.14275},
  year={2025}
}

@article{gao2024omni,
  title={Omni-math: A universal olympiad level mathematic benchmark for large language models},
  author={Gao, Bofei and Song, Feifan and Yang, Zhe and Cai, Zefan and Miao, Yibo and Dong, Qingxiu and Li, Lei and Ma, Chenghao and Chen, Liang and Xu, Runxin and others},
  journal={arXiv preprint arXiv:2410.07985},
  year={2024}
}

@book{aczel2013homotopy,
  title={Homotopy type theory: univalent foundations of mathematics},
  author={Aczel, Peter and Ahrens, Benedikt and Altenkirch, Thorsten and Awodey, Steve and Barras, Bruno and Bauer, Andrej and Bertot, Yves and Bezem, Marc and Coquand, Thierry and Finster, Eric and others},
  year={2013},
  publisher={The Univalent Foundations Program Institute for Advanced Study}
}

@inproceedings{papineni2002bleu,
  title={Bleu: a method for automatic evaluation of machine translation},
  author={Papineni, Kishore and Roukos, Salim and Ward, Todd and Zhu, Wei-Jing},
  booktitle={Proceedings of the 40th annual meeting of the Association for Computational Linguistics},
  pages={311--318},
  year={2002}
}

@inproceedings{wang2018first,
  title={First experiments with neural translation of informal to formal mathematics},
  author={Wang, Qingxiang and Kaliszyk, Cezary and Urban, Josef},
  booktitle={International Conference on Intelligent Computer Mathematics},
  pages={255--270},
  year={2018},
  organization={Springer}
}

@inproceedings{sidhu2024evaluation,
  title={An Evaluation Benchmark for Autoformalization in Lean4},
  author={Sidhu, Jasdeep and Mishra, Shubhra and Gulati, Aryan and Ladsaria, Devanshu and Miranda, Brando},
  booktitle={The Second Tiny Papers Track at ICLR 2024}
}

@article{xiong2020approximate,
  title={Approximate nearest neighbor negative contrastive learning for dense text retrieval},
  author={Xiong, Lee and Xiong, Chenyan and Li, Ye and Tang, Kwok-Fung and Liu, Jialin and Bennett, Paul and Ahmed, Junaid and Overwijk, Arnold},
  journal={arXiv preprint arXiv:2007.00808},
  year={2020}
}

@article{gao2023retrieval,
  title={Retrieval-augmented generation for large language models: A survey},
  author={Gao, Yunfan and Xiong, Yun and Gao, Xinyu and Jia, Kangxiang and Pan, Jinliu and Bi, Yuxi and Dai, Yixin and Sun, Jiawei and Wang, Haofen and Wang, Haofen},
  journal={arXiv preprint arXiv:2312.10997},
  volume={2},
  number={1},
  year={2023}
}

@article{asl2024robustsentembed,
  title={Robustsentembed: Robust sentence embeddings using adversarial self-supervised contrastive learning},
  author={Asl, Javad Rafiei and Panzade, Prajwal and Blanco, Eduardo and Takabi, Daniel and Cai, Zhipeng},
  journal={arXiv preprint arXiv:2403.11082},
  year={2024}
}

@inproceedings{lee2022deduplicating,
  title={Deduplicating training data makes language models better},
  author={Lee, Katherine and Ippolito, Daphne and Nystrom, Andrew and Zhang, Chiyuan and Eck, Douglas and Callison-Burch, Chris and Carlini, Nicholas},
  booktitle={Proceedings of the 60th Annual Meeting of the Association for Computational Linguistics (Volume 1: Long Papers)},
  pages={8424--8445},
  year={2022}
}

@article{karkkainen2006linear,
  title={Linear work suffix array construction},
  author={K{\"a}rkk{\"a}inen, Juha and Sanders, Peter and Burkhardt, Stefan},
  journal={Journal of the ACM (JACM)},
  volume={53},
  number={6},
  pages={918--936},
  year={2006},
  publisher={ACM New York, NY, USA}
}

@article{liu2024infini,
  title={Infini-gram: Scaling unbounded n-gram language models to a trillion tokens},
  author={Liu, Jiacheng and Min, Sewon and Zettlemoyer, Luke and Choi, Yejin and Hajishirzi, Hannaneh},
  journal={arXiv preprint arXiv:2401.17377},
  year={2024}
}

@article{ying2024internlm,
  title={Internlm-math: Open math large language models toward verifiable reasoning},
  author={Ying, Huaiyuan and Zhang, Shuo and Li, Linyang and Zhou, Zhejian and Shao, Yunfan and Fei, Zhaoye and Ma, Yichuan and Hong, Jiawei and Liu, Kuikun and Wang, Ziyi and others},
  journal={arXiv preprint arXiv:2402.06332},
  year={2024}
}

@article{guo2025deepseek,
  title={Deepseek-r1 incentivizes reasoning in llms through reinforcement learning},
  author={Guo, Daya and Yang, Dejian and Zhang, Haowei and Song, Junxiao and Wang, Peiyi and Zhu, Qihao and Xu, Runxin and Zhang, Ruoyu and Ma, Shirong and Bi, Xiao and others},
  journal={Nature},
  volume={645},
  number={8081},
  pages={633--638},
  year={2025},
  publisher={Nature Publishing Group UK London}
}

@article{ai2025ming,
  title={Ming-Flash-Omni: A Sparse, Unified Architecture for Multimodal Perception and Generation},
  author={AI, Inclusion and Ma, Bowen and Zou, Cheng and Yan, Canxiang and Jin, Chunxiang and Shen, Chunjie and Zheng, Dandan and Wang, Fudong and Xu, Furong and Yao, GuangMing and others},
  journal={arXiv preprint arXiv:2510.24821},
  year={2025}
}

@techreport{anthropic_claude35_sonnet_2024,
  author       = {Anthropic},
  title        = {Claude 3.5 Sonnet},
  institution  = {Anthropic},
  year         = {2024},
  url          = {https://www.anthropic.com/news/3-5-models-and-computer-use},
}

@article{achiam2023gpt,
  title={Gpt-4 technical report},
  author={Achiam, Josh and Adler, Steven and Agarwal, Sandhini and Ahmad, Lama and Akkaya, Ilge and Aleman, Florencia Leoni and Almeida, Diogo and Altenschmidt, Janko and Altman, Sam and Anadkat, Shyamal and others},
  journal={arXiv preprint arXiv:2303.08774},
  year={2023}
}

@misc{qwen2025qwen25technicalreport,
      title={Qwen2.5 Technical Report}, 
      author={Qwen and Yang , An and Yang , Baosong and Zhang , Beichen and Hui , Binyuan and Zheng , Bo and others},
      year={2025},
      eprint={2412.15115},
      archivePrefix={arXiv},
      primaryClass={cs.CL},
      url={https://arxiv.org/abs/2412.15115}, 
}

@article{yang2025qwen3,
  title={Qwen3 technical report},
  author={Yang, An and Li, Anfeng and Yang, Baosong and Zhang, Beichen and Hui, Binyuan and Zheng, Bo and Yu, Bowen and Gao, Chang and Huang, Chengen and Lv, Chenxu and others},
  journal={arXiv preprint arXiv:2505.09388},
  year={2025}
}

@article{jiang2023multilingual,
  title={Multilingual mathematical autoformalization},
  author={Jiang, Albert Q and Li, Wenda and Jamnik, Mateja},
  journal={arXiv preprint arXiv:2311.03755},
  year={2023}
}

@inproceedings{zhao2025swift,
  title={Swift: a scalable lightweight infrastructure for fine-tuning},
  author={Zhao, Yuze and Huang, Jintao and Hu, Jinghan and Wang, Xingjun and Mao, Yunlin and Zhang, Daoze and Jiang, Zeyinzi and Wu, Zhikai and Ai, Baole and Wang, Ang and others},
  booktitle={Proceedings of the AAAI Conference on Artificial Intelligence},
  volume={39},
  number={28},
  pages={29733--29735},
  year={2025}
}

\newpage
\appendix

\section{Additional Results}

\subsection{Full Figures}
\begin{figure*}
	\centering
	\includegraphics[width=1.0\textwidth]{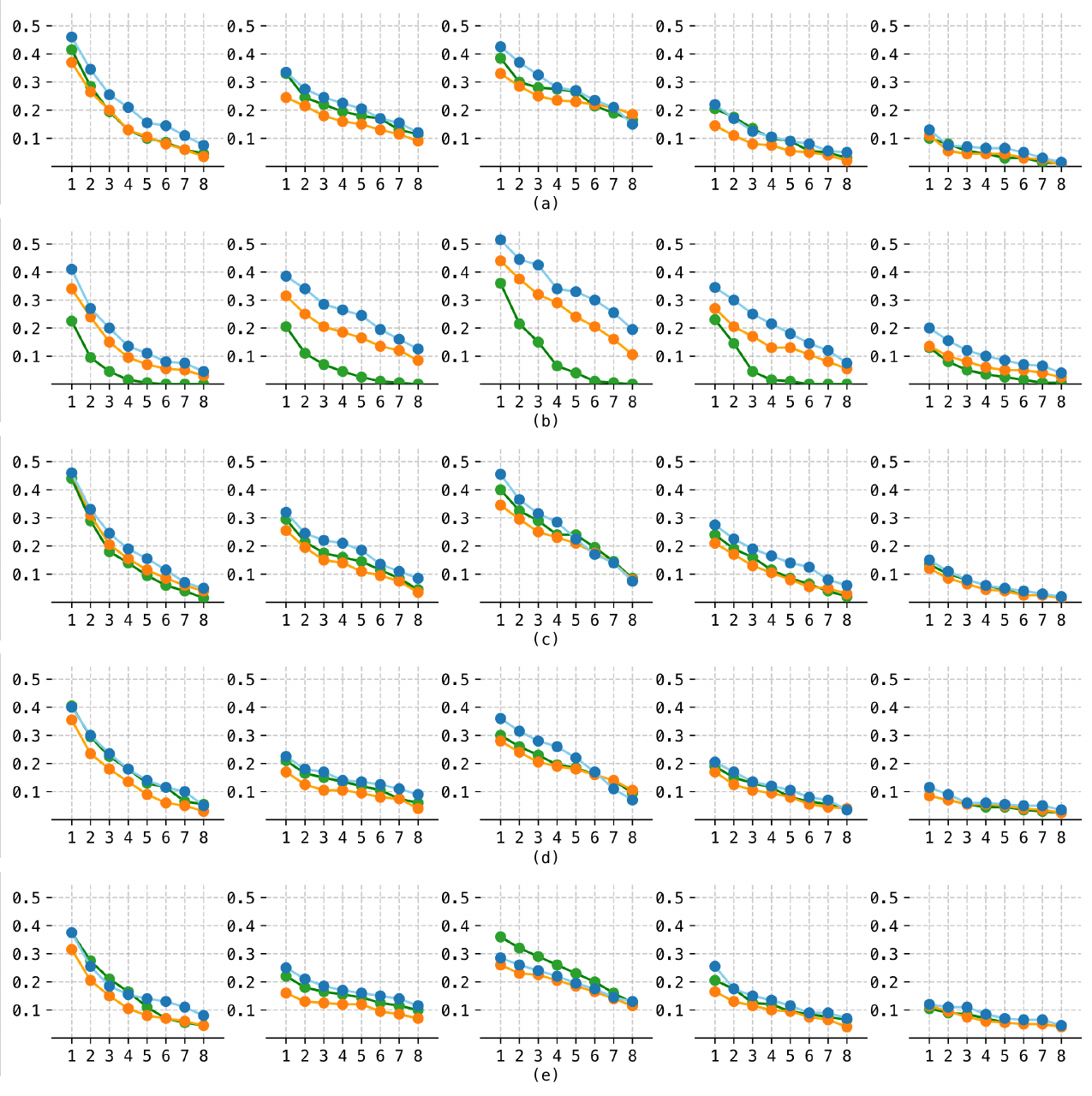}
	\caption{LLM consistency across multiple attempts. From \textbf{left} to \textbf{right} correspond to Diff01, Diff23, Diff45, Diff67, and Diff89; from \textbf{top} to \textbf{bottom} correspond to Claude-3.5-Sonnet, DeepSeek-R1, GPT-4o, Ling-flash-2.0, and Qwen-Max. Each subplot illustrates the proportion of problems ($y$ axis) for which at least k (from 1 to 8, $x$ axis) out of 8 attempts successfully pass the BEq verification.
		Dependency retrieval method:
		\textcolor{blue}{DDR (blue)}; \textcolor{orange}{ICL (orange)}; \textcolor{lightgreen}{N/A (green)}.
	}
	\label{fig:example-all_main2}
\end{figure*}

\begin{figure*}
	\centering
	\includegraphics[width=1.0\textwidth]{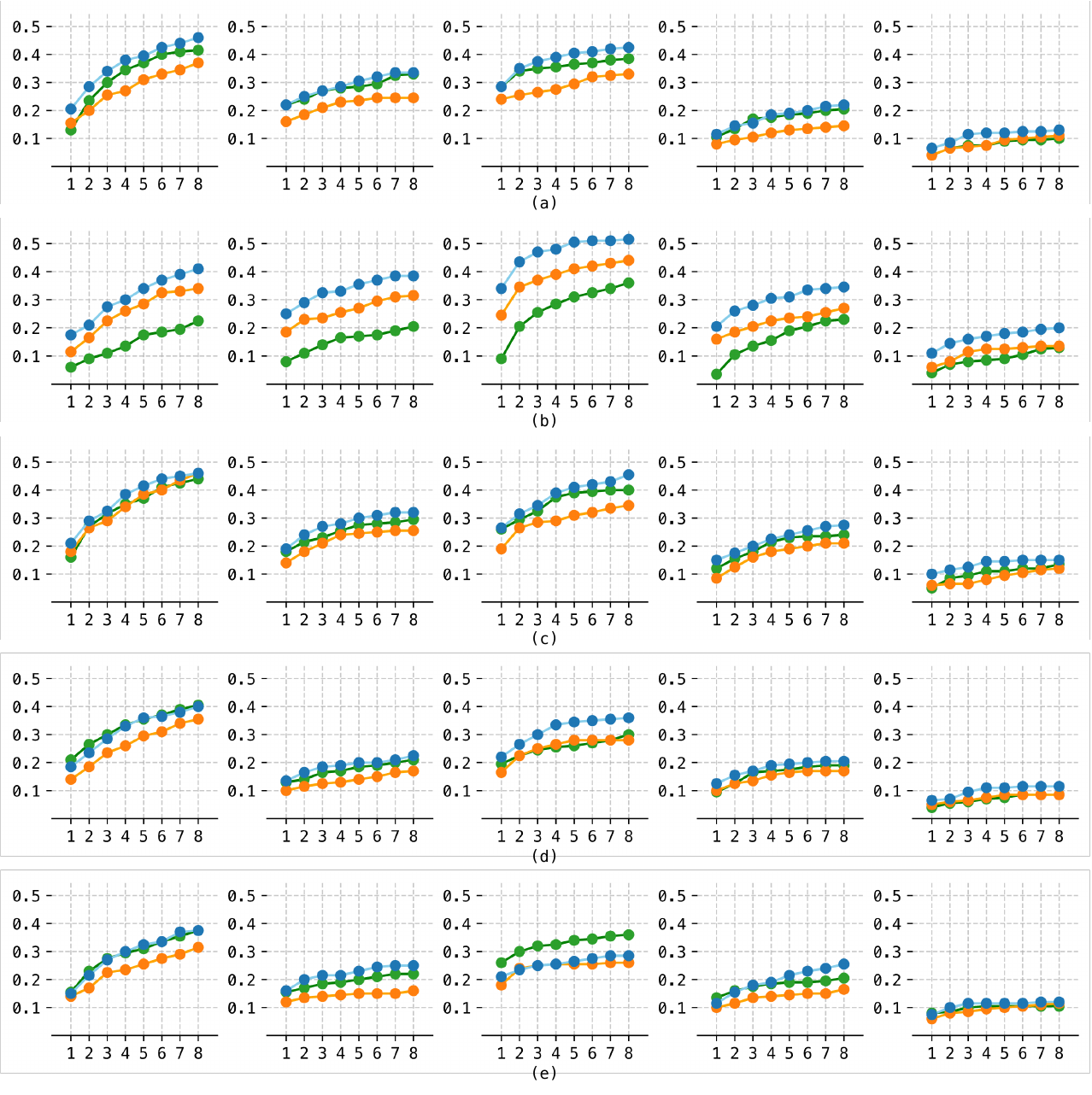}
	\caption{LLM efficiency across multiple attempts. From \textbf{left} to \textbf{right} correspond to Diff01, Diff23, Diff45, Diff67, and Diff89; from \textbf{top} to \textbf{bottom} correspond to Claude-3.5-Sonnet, DeepSeek-R1, GPT-4o, Ling-flash-2.0, and Qwen-Max. Each subplot illustrates the intermediate pass@k score ($y$ axis) out of 8 attempts using BEq verification, k range from 1 to 8 ($x$ axis).
		Dependency retrieval method:
		\textcolor{blue}{DDR (blue)}; \textcolor{orange}{ICL (orange)}; \textcolor{lightgreen}{N/A (green)}.
	}
	\label{fig:example-all_main3}
\end{figure*}

While Figure \ref{fig:example-deepseek_r1_main2} and Figure \ref{fig:example-deepseek_r1_main3} primarily presented the experimental results on the Deepseek-R1 model, this section provides a more comprehensive evaluation by integrating the DDR method with various other ICL methods in Figure \ref{fig:example-all_main2} and Figure \ref{fig:example-all_main3}. The experimental data demonstrates that incorporating DDR brings significant performance improvements to these baseline methods in the vast majority of cases.

\subsection{Out of Domain Study}
\label{sec:out-of-domain}
To evaluate the generalization ability of DDR, we conduct autoformalization experiments on the Omni-math dataset \citep{gao2024omni}, adopting the experimental setup from \citet{lin2025goedel}. We compare DDR against the baseline methods Goedel-Formalizer-V2-8B \citep{lin2025goedel} and Kimina-Autoformalizer-7B \citep{kimina_prover_2025}. 
To evaluate the correctness of the generated formalizations, we employ the method proposed in \citet{lin2025goedel}, which combines Lean4 TypeC with semantic judgments from LLMs. The pass@8 results are presented in Table \ref{tab:out-of-domain-0}. 
The results demonstrate that DDR enhances the performance of existing autoformalizers, even those not specifically designed for dependency retrieval.

\begin{table}
	\centering
	\begin{adjustbox}{width=.5\textwidth,center}
		\begin{tabular}{l|ccc}
			\toprule
			\textbf{Name} &
			\multicolumn{1}{c}{\textbf{RAG}} &
            \multicolumn{1}{c}{\textbf{Pass}} &
			\multicolumn{1}{c}{\textbf{Failed}}
            \\
			
			\midrule
			Goedel-Formalizer-V2-8B & N/A & 180 & 120\\
           Goedel-Formalizer-V2-8B & DDR & \textbf{183} & \textbf{117}\\[2pt]
			\hdashline \\ [-10pt]
           Kimina-Autoformalizer-7B  & N/A & 105 & 195\\
           Kimina-Autoformalizer-7B  & DDR & \textbf{128} & \textbf{172}\\
			\bottomrule
		\end{tabular}
	\end{adjustbox}
	\caption{Comparison of different formalizers on 300 Omni-math problems.}
	\label{tab:out-of-domain-0}
\end{table}

\section{Examples Illustration of Diff01}
\label{append:examples-illustration}

A notable observation is that samples categorized as {difficulty} 0 and {difficulty} 1 present a counter-intuitively high formalization challenge. 
This is primarily because the natural language statements in this category often do not represent abstract descriptions of the problem's mathematical essence, as illustrated by examples in Figure \ref{fig:case_study_1}. 
Our subsequent experimental results corroborate this finding: for \textbf{Diff01} dataset, the performance of autoformalization across all methods are worse than \textbf{Diff23} and \textbf{Diff45}, yet still better than that of \textbf{Diff89}. 

Figure \ref{fig:case_study_1} presents several examples from the \textbf{Diff01} dataset. These problems are primarily formulated in a natural language style rather than as rigorous formal mathematical statements. 
Such presentation brings additional challenge in autoformalization, and renders difficulty metrics that rely on \textbf{Depend Rate} and \textbf{Depend Length} inapplicable, thereby impeding an accurate quantification of the autoformalization task difficulty on this dataset. 
Furthermore, because these problem descriptions lack an abstract representation of their mathematical essence, all methods evaluated in Table \ref{tab:results-3} and Table \ref{tab:results-icl} exhibit a significant degradation in performance on the \textbf{Diff01} dataset.

\begin{figure*}
\definecolor{burntorange}{rgb}{0.8, 0.33, 0.0}
\begin{tcolorbox}[colback=mybrown!5!white,colframe=mybrown!75!black]
\begin{small}

\textbf{Informal statement example 1:}\\
Prove that the total number of wooden blocks Thomas used to make the five stacks is 55.\\

\vspace{-0.6em}
\textbf{Formal statement example 1:}
\vspace{-1.9em}
\begin{lstlisting}[style=isabelle]
theorem thm_P
(a : $\mathbb{N}$) ($\text{h}_0$ : a > 0) ($\text{h}_1$ : a + (a + 1) + (a + 2) + (a + 3) + (a + 4) = 55) :
a = 5 := by sorry
\end{lstlisting}
 
\vspace{1.0em}

\textbf{Informal statement example 2:}\\
Prove that the time it takes for Usain to run a whole lap around the school stadium is 42\\

\vspace{-0.6em}
\textbf{Formal statement example 2:}
\vspace{-1.9em}
\begin{lstlisting}[style=isabelle]
theorem thm_P :
$\exists$ t, t > 0 $\wedge$ t = 42 := by sorry
\end{lstlisting}

\vspace{1.0em}

\textbf{Informal statement example 3:}\\
Prove that the number of ways $n$ tourists can distribute themselves among $n$ cinemas so that each cinema has exactly one tourist is $n!$.\\

\vspace{-0.6em}
\textbf{Formal statement example 3:}
\vspace{-1.9em}
\begin{lstlisting}[style=isabelle]
theorem thm_P
(n : $\mathbb{N}$) : {f : Fin n $\rightarrow$ Fin n | f.Injective}.ncard = n! := by sorry
\end{lstlisting}

\vspace{1.0em}

\textbf{Informal statement example 4:}\\
Prove that the probability that James wins when he and his sister each spin a modified spinner with six congruent sectors numbered from 1 to 6, and James wins if the absolute difference of their numbers is 2 or less, is $\frac{2}{3}$.\\

\vspace{-0.6em}
\textbf{Formal statement example 4:}
\vspace{-1.9em}
\begin{lstlisting}[style=isabelle]
theorem thm_P :
(Set.ncard {x : Fin 6 $\times$ Fin 6 | x.1 - x.2 $\in$ Finset.Icc (-2) 2}) / 36 
= 2 / 3 := by sorry
\end{lstlisting}

\end{small}
\vspace{-0.7em}
\end{tcolorbox}
\vspace{-0.7em}
\caption{Examples from the FineLeanCorpus with \textit{difficulty} levels 0 and 1. }
\label{fig:case_study_1}
\end{figure*}

\section{Experiment Settings}
\label{append:experiment-details}

\subsection{SAC Details}
We employ precision and recall as quantitative metrics to evaluate the performance of the dependency retrieval results. 
A key challenge in this evaluation arises from the discrepancy between the outputs of DDR, which directly generate dependency short names, and the ground-truth labels, which are Mathlib object identifiers extracted from source code. 
This leads to name ambiguity, as source code often contains unqualified or partially qualified names. 
To accurately identify true positives for our evaluation, we introduce a suffix-based matching criterion. 
Specifically, we represent each identifier as a sequence of dot-separated components. A retrieved identifier is considered a match for a ground-truth label if and only if two conditions are met: (1) the string representation of one identifier is a substring of the other; and (2) when both identifiers are split by the dot character (\verb|'|.\verb|'|) into lists of their components, one list is a sublist of the other.

\subsection{Metrics}
To evaluate the semantic equivalence between the formal statements predicted by the model and the ground-truth formal statements, we employ two criteria: Type Checking (TypeC) and Bidirectional Extended Definitional Equality (BEq \citep{liu2025rethinking}. TypeC ensures the syntactic correctness of expressions, whereas BEq is utilized to determine their semantic equivalence. 
Building upon these criteria, we define the TypeC@8 and BEq@8 as our final evaluation metrics shown in Table \ref{tab:results-3} and Table \ref{tab:results-icl}. 
These metrics measure the percentage of problems for which at least one of the top-8 predicted statements is deemed equivalent to the ground-truth statement under the corresponding criterion.

\subsection{Hallucination Details}

It is important to note that the hallucination metric, \textbf{Hall}, is computed exclusively over the set of {effective samples}. We define an effective sample as an instance for which the model retrieves one or more dependencies from the library. Some autoformalization tasks are simple enough to be solved using only the given context and basic syntax, thereby not requiring any explicit library dependencies. For such cases, our model is designed to output a predefined token, \verb|"|None\verb|"| (or \verb|"|No usage\verb|"| in our implementation), to indicate that no dependency retrieval is necessary. Since these instances do not involve an actual retrieval action, the concept of hallucination, retrieving an incorrect or non-existent item, is not applicable. Therefore, they are excluded from the metric's calculation. For example, in a dataset of 200 retrieved dependency samples, if 50 are resolved with a \verb|"|None\verb|"| output, the effective sample size for computing the mean hallucination rate (\textbf{Hall$_\text{m}$}) and its standard deviation (\textbf{Hall$_\text{std}$}) is 150.

\subsection{Baseline Details}
To comprehensively evaluate our proposed method, we established several baseline models and conducted ablation studies. For the ablation studies, we employed the RAutoformalizer (RA) from \citet{liu2025rethinking} as a pure formalizer baseline, and RA+R@k as an autoformalizer that incorporates the top-k retrieved dependencies using the method in \citet{liu2025rethinking}. The RA was fine-tuned based on InternLM2-Math-Base-7B \citep{ying2024internlm}. Our baselines cover multiple aspects: first, we evaluated a range of state-of-the-art general LLMs via ICL, including DeepSeek-R1 \citep{guo2025deepseek}, Ling-flash-2.0 \citep{ai2025ming}, Claude-3.5-Sonnet (claude-3-5-sonnet-20241022) \citep{anthropic_claude35_sonnet_2024}, GPT-4o (gpt-4o-2024-08-06) \citep{hurst2024gpt}, and Qwen-Max (qwen-max-2024-09-19) \citep{team2024qwen2}; second, we included an InternLM2-Math-Base-7B \citep{ying2024internlm} model fine-tuned on the MMA Lean subset \citep{jiang2023multilingual}. For our proposed DDR method, we selected Qwen3-32B \citep{yang2025qwen3} as its base model. This choice highlights a core advantage of DDR: its flexibility in selecting a base dependency retrieval model without being restricted to specific embedding models.

During the decoding stage for all experiments, the temperature was set to 0.7, and we generated 8 samples for each problem. 
To ensure a fair comparison, the dependency retrieval step for the DDR method was performed only once, and its results were reused for all subsequent Lean 4 statement formalization generations. It is noteworthy that none of the autoformalization models discussed in the main text, including the fine-tuned ones, have undergone SFT on the FineLeanCorpus \citep{peng2025criticlean}. 

For a fair comparison, we adopt an experimental setup identical to that of \citet{liu2025rethinking}. We used the same prompt as described in Section A.7.1 of \citet{liu2025rethinking} for evaluation on BEq. Similarly, the prompt for the ICL method is also consistent with that in Section A.7.3 of \citet{liu2025rethinking}. 
Gemini 2.5 pro is utilized as an auxiliary tool for BEq. 
Gemini 2.5 pro is also the LLM judge used in section \ref{sec:out-of-domain}.

Our proposed DDR method is fine-tuned using the Swift \citep{zhao2025swift} framework with the following hyperparameters:
\begin{enumerate}[label=\textbullet]
	\item \quad Max Sequence Length: 8192
	\item \quad Batch size: 1
	\item \quad Gradient Accumulation: 64
	\item \quad Training Devices: 8
	\item \quad Train Epochs: 6
	\item \quad Optimizer: AdamW with learning rate \(1 \times 10^{-4}\), \(\beta = (0.9, 0.95)\), weight decay 0.1, maximal gradient norm 1.0, no warmup.
	\item \quad Learning Rate Scheduler: Cosine.
	\item \quad LoRA: rank 32, alpha 32, dropout 0.05.
\end{enumerate}

\section{Prompt Templates}
\subsection{Prompt Template for Dependency Retrieval}

\texttt{
You are an expert Lean 4 `Mathlib` dependency extractor.}

\texttt{Your task is to analyze a given mathematical proposition and identify the key mathematical concepts that would require specific definitions, theorems, or notations from the Lean 4 `Mathlib` library.}

\texttt{**CRITICAL INSTRUCTIONS:**}

\texttt{1.  Your output **MUST** be a single, valid JSON object.}

\texttt{2.  The JSON object **MUST** have one key: "dependency".}

\texttt{3.  The value for the "dependency" key **MUST** be a list of strings. Each string is a potential `Mathlib` dependency.}

\texttt{4.  The dependency names can be fully qualified (e.g., `Nat.minFac`) or common shortcuts (e.g., `ncard`, `Icc`, `MOD`).}

\texttt{5.  If there is no potential `Mathlib` library dependency, output \{"dependency":["No usage"]\}}

\texttt{6.  Do **NOT** write any Lean code. Do **NOT** provide any explanations or text outside of the JSON object. Your entire response must be only the JSON.}

\texttt{
Here are some examples of the required input and output:}\\

\noindent
\texttt{**Example 1**}

\noindent
\texttt{**Math Proposition:**}

\noindent
\texttt{Prove that the remainder when the number N of ordered 2011 - tuples of positive integers (a\_1, a\_2, \textbackslash ldots, a\_\{2011\}) with 1 \textbackslash le a\_1, a\_2, \textbackslash ldots, a\_\{2011\} \textbackslash le 2011\textasciicircum 2 such that there exists a polynomial f of degree 4019 satisfying the following conditions:}

\noindent
\texttt{1. f(n) is an integer for every integer n;}

\noindent
\texttt{2. 2011\textasciicircum 2 \textbackslash mid f(i) - a\_i for i = 1, 2, \textbackslash ldots, 2011;}

\noindent
\texttt{3. 2011\textasciicircum 2 \textbackslash mid f(n + 2011) - f(n) for every integer n;}

\noindent
\texttt{is divided by 1000 is equal to 211.}

\noindent
\texttt{**Output:**}

\noindent
\verb|```|
\texttt{
\{}

\noindent
\texttt{dependency:}

\noindent
\texttt{[Polynomial,degree,Fin,eval,Set.ncard]}

\noindent
\texttt{\}}

\noindent
\verb|```|

\quad \\

\noindent
\texttt{**Example 2**}

\noindent
\texttt{**Math Proposition:**}

\noindent
\texttt{Prove that the set of integers (n \textbackslash ge 1) for which the smallest prime divisor of ((n!)\textasciicircum n + 1) is at most (n + 2015) is finite.}

\noindent
\texttt{**Output:**}

\noindent
\verb|```|
\texttt{\{}

\noindent
\texttt{dependency:[Nat.factorial,Finite,}

\noindent
\texttt{Nat.minFac]}

\noindent
\texttt{
\}}

\noindent
\verb|```|

\quad\\

\noindent
\texttt{**Example 3**}

\noindent
\texttt{**Math Proposition:**}

\noindent
\texttt{Prove that for a positive odd integer n, the arithmetic mean of the fractional parts \textbackslash left\{\textbackslash frac\{k\textasciicircum\{2n\}\}\{p\}\textbackslash right\} for k = 1, \textbackslash ldots, \textbackslash frac\{p - 1\}\{2\} is \textbackslash frac\{1\}\{2\} for infinitely many primes p satisfying p \textbackslash equiv 2 \textbackslash pmod\{n\} and p \textbackslash equiv 1 \textbackslash pmod\{4\}.}

\noindent
\texttt{**Output:**}

\noindent
\verb|```|
\texttt{
\{}

\noindent
\texttt{dependency:[MOD,Nat.Prime,Odd,Finset.Icc]}

\noindent
\texttt{
\}}

\noindent
\verb|```|

\quad\\

\noindent
\texttt{**Example 4**}

\noindent
\texttt{**Math Proposition:**}

\noindent
\texttt{Prove that if t represents the number ten, then the value of the expression (t + t + t + t) is 40.}

\noindent
\texttt{**Output:**}

\noindent
\verb|```|
\texttt{
\{}

\noindent
\texttt{dependency:[No usage]}

\noindent
\texttt{
\}}

\noindent
\verb|```|

\noindent
\texttt{Now, perform this task for the following proposition.}

\noindent
\texttt{**Math Proposition:**}

\noindent
\texttt{\{input\}}

\noindent
\texttt{**Output:**}

\noindent
\verb|```|

\quad\\

\noindent
\verb|```|

\subsection{Prompt Template for Autoformalization with Dependency Retrieval}

\noindent
\texttt{Please translate mathematical propositions into Lean 4 theorems. `Mathlib` is the MUST import.}

\noindent
\texttt{
DO NOT try to prove the theorem, ONLY translate it.}

\noindent
\texttt{
Use `sorry` as placeholder.
Use the given potential `Mathlib` dependency items.}

\noindent
\texttt{
Here are some examples:}

\quad\\

\noindent
\texttt{Math Proposition:}

\noindent
\texttt{Suppose f : R \textbackslash rightarrow R is continuous and for every \textbackslash alpha>0, \textbackslash lim\_\{n\textbackslash rightarrow \textbackslash infty\}f(n\textbackslash alpha) = 0.}

\noindent
\texttt{Prove that
\textbackslash lim\_\{x\textbackslash rightarrow \textbackslash infty\}f(x) = 0.
The potential `Mathlib` dependency are }

\noindent
\texttt{[Real,Continuous,Tendsto,Nat,atTop].}

\noindent
\texttt{Lean Theorem:}

\noindent
\verb|```|

\noindent
\texttt{import Mathlib}

\noindent
\texttt{theorem exercise}

\noindent
\texttt{(f : Real \textbackslash r  Real)}

\noindent
\texttt{(hf : Continuous f \textbackslash and  \textbackslash all \textbackslash alpha > 0, Tendsto (fun n : Nat \textbackslash ma f (n * \textbackslash alpha)) atTop (\textbackslash MCN 0)):}

\noindent
\texttt{(Tendsto f atTop (\textbackslash MCN 0)) :=
sorry}

\noindent
\verb|```|

\quad\\

\noindent
\texttt{Math Proposition:}

\noindent
\texttt{Let G be a finite group (with a multiplicative operation), and A be a subset of G
that contains more than half of G's elements. Prove that every element of G can
be expressed as the product of two elements of A.
The potential `Mathlib` dependency are}
 
\noindent
\texttt{[Group,Finite,Set,ncard,Nat.card,Rat].}

\noindent
\texttt{Lean Theorem:}

\noindent
\verb|```|

\noindent
\texttt{import Mathlib}

\noindent
\texttt{open Filter Topology}

\noindent
\texttt{theorem exercise}

\noindent
\texttt{[Group G]}

\noindent
\texttt{(hG : Finite G)}

\noindent
\texttt{(A : Set G)}

\noindent
\texttt{(hA : A.ncard > (Nat.card G : Rat)/2):}

\noindent
\texttt{\textbackslash all g : G, \textbackslash ex x \textbackslash in  A, \textbackslash ex y \textbackslash in  A, g = x * y :=
sorry}

\noindent
\verb|```|

\quad\\

\noindent
\texttt{Math Proposition:}

\noindent
\texttt{Let p be a prime number greater than 5. Let f(p) denote the number of infinite
sequences a\_1,a\_2,a\_3,... such that a\_n \textbackslash in \{1,2,...,p-1\} and a\_\{n\}a\_\{n+2\} \textbackslash equiv 1 + a\_\{n+1\} (mod p)
for all n \textbackslash ge 1. Prove that f(p) is congruent to 0 or 2 (mod 5).
The potential `Mathlib` dependency are }

\noindent
\texttt{[Nat,Nat.Prime,ZMod,ncard,MOD].}

\noindent
\texttt{Lean Theorem:}

\noindent
\verb|```|

\noindent
\texttt{import Mathlib}

\noindent
\texttt{theorem exercise}

\noindent
\texttt{(p : Nat)}

\noindent
\texttt{(hp : Nat.Prime p \textbackslash and  p > 5)}

\noindent
\texttt{(f : Nat := \{a : Nat \textbackslash r (ZMod p) | \textbackslash all n : Nat, a n \textbackslash neq  0 \textbackslash and  a n * a (n + 2) = 1 + a (n + 1)\}.ncard):}

\noindent
\texttt{f \textbackslash == 0 [MOD 5] \textbackslash or  f \textbackslash == 2 [MOD 5] :=
sorry}

\noindent
\verb|```|

\quad\\

\noindent
\texttt{Math Proposition:}

\noindent
\texttt{Let p be 10, and q be 15, prove that p+b equals 25.
The potential `Mathlib` dependency are}

\noindent
\texttt{[No usage].}

\noindent
\texttt{Lean Theorem:}

\verb|```|

\noindent
\texttt{import Mathlib}

\noindent
\texttt{theorem exercise}

\noindent
\texttt{(p : Nat)}

\noindent
\texttt{(q : Nat)}

\noindent
\texttt{(hp : p = 10)}

\noindent
\texttt{(hq : q = 15):}

\noindent
\texttt{p + q = 25 :=
sorry}

\noindent
\verb|```|

\quad\\

\noindent
\texttt{Please translate the following proposition:}

\noindent
\texttt{Math Proposition:}

\noindent
\texttt{\{input\}}

\noindent
\texttt{The potential `Mathlib` dependency are}
 
\noindent
\texttt{\{retrieved-dependency\}}

\noindent
\texttt{Lean Theorem:}
\noindent
\verb|```|

\quad\\

\noindent
\verb|```|

\subsection{Other Templates}

Other ICL templates employed in this paper, including those for autoformalization without dependency retrieval and BEq verification, adhere to the corresponding settings in \citet{liu2025rethinking}.

\end{document}